\documentclass[11pt]{article}

\usepackage[T1]{fontenc}
\usepackage[utf8]{inputenc}
\usepackage{lmodern}
\usepackage{microtype}
\usepackage{graphicx}
\usepackage{amsmath,amssymb}
\usepackage{tabularx}
\usepackage{booktabs}
\usepackage{array}
\usepackage{multirow}
\usepackage{siunitx}
\usepackage{subcaption}
\usepackage{float}
\usepackage[numbers,sort&compress]{natbib}
\usepackage[hidelinks]{hyperref}

\title{A New Generalised Three-Dimensional Failure Criterion for In-Situ Stress Conditions}
\author{
Riu Riu Baglary\textsuperscript{1}
\and
Jwngsar Brahma\textsuperscript{1,*}\\[2pt]
\small \textsuperscript{1}Department of Applied Sciences, Tezpur University, Assam, India\\
\small \textsuperscript{*}Corresponding author: \href{mailto:jwngsar@gmail.com}{jwngsar@gmail.com}\\
\small Riu Riu Baglary: \href{mailto:baglaryriu@gmail.com}{baglaryriu@gmail.com}
}
\date{}

\begin{document}

\maketitle
\begin{abstract}

Three-dimensional failure criteria are important for predicting rock failure under in-situ stresses, particularly in wellbore stability and sand-production applications. This study proposes a generalized three-dimensional failure criterion that expresses the major principal stress at failure as a linear function of the intermediate and minor principal stresses. Unlike formulations that incorporate the intermediate principal stress through a weighted mean stress or mean average while assigning equal weights to the major and minor stresses, the proposed criterion introduces its contribution into the effective normal stress through unequal weighting coefficients. Derived from the Mohr–Coulomb criterion, the formulation reduces to the conventional Mohr–Coulomb criterion when the intermediate-stress weighting parameter is zero. Triaxial compression data for eleven rock types are used to calibrate and evaluate the model. The calibrated weighting parameter K ranges from 0.03 to 0.38, indicating variations in the contribution of the intermediate principal stress. The criterion is assessed in Haigh–Westergaard coordinates and compared with the New Extended Mohr–Coulomb, Mogi–Coulomb, and Drucker–Prager criteria using the root mean square error (RMSE) and coefficient of determination (DC). The proposed criterion produces RMSE values of 5.13–36.62 MPa and DC values of 0.88–0.98. It yields lower RMSE values than the New Extended Mohr–Coulomb and Drucker–Prager criteria for all eleven rock types, while showing comparable or improved performance relative to the Mogi–Coulomb criterion, depending on the rock type. For the illustrative parameters examined, its deviatoric-plane failure envelope is smaller, indicating a more conservative strength prediction. The proposed criterion provides a framework for representing three-dimensional rock failure and has application in wellbore-stability assessment.

\end{abstract}
\section{Introduction}
The prediction of rock failure in three-dimensional stress states is a fundamental geomechanics topic that has direct implications for wellbore stability and other subsurface engineering applications. The increasing demand for hydrocarbons extraction from deeper reservoirs and increasingly challenging geological environments has made wellbore stability a critical issue in today’s oil and gas industry. As drilling depth increases beyond 4 km, over burden pressure, tectonic effect, high temperature and rock’s anisotropy plays critical roles \cite{kim1984modelling}. To mitigate the instability in drilling operation a new technique and tools are essentials. The rock failure criteria plays an important role to develop such technique and tools.
\par Rock breakout from compressive failure and rock fracturing from tensile failure are the two main types of wellbore instability. These can result from chemical interactions between the rock, notably clay, and the drilling fluid, or from mechanical effects like high stresses, insufficient rock strength, or poor drilling techniques. Prior to drilling, the rock exists in a stress state of equilibrium, characterized by in-situ stresses or far field stresses: vertical stress(overburden stress), maximum horizontal stress, and minimum horizontal ($\sigma_V , \sigma_H, \sigma_h$) \cite{gaurina1994mechanical}.  Failure criteria of rock provide necessary tools to observe and predict how rocks will response mechanically under different loading conditions. 
\par 
The Mohr-Coulomb criterion is one of the simplest and widely used strength criteria in rock engineering due to the clear physical meaning of its parameters, and it is based on two key assumptions. First, during the failure, the major principal stress ($\sigma_1$) increases linearly with the minor principal stress ($\sigma_3$). Secondly, the intermediate principal stress ($\sigma_2$) has no effect on rock strength.  Hoek–Brown (\cite{hoek1997practical,hoek2002hoek}) failure
criterion is an empirically derived nonlinear relationship that characterizes the failure condition of both intact and jointed rock masses. It relates the major ($\sigma_1$) and minor ($\sigma_3$) principal stresses at failure but neglects the influence of the intermediate principal
stress ($\sigma_2$). However, much evidence has emerged suggesting that the intermediate principal stress significantly influence the rock strength (\cite{haimson2009three,mogi1971fracture,takahashi1989effect}). 
\par 
Numerous three-dimensional failure criteria have been developed to address the shortcomings of conventional two-dimensional models, including the Mohr-Coulomb criterion, which overlooks the effect of the intermediate principal stress ($\sigma_2$). Criteria such as those proposed by Drucker and Prager \cite{drucker1952soil}, Lade criterion \cite{kim1984modelling}, Paul criterion \cite{paul1968generalized}, Meyer and Labuz \cite{meyer2013linear}, Jiang (\cite{jiang2015failure,jiang2018simple}), as well as the Generalized Mohr-Culomb \cite{tian2023generalized} and many more have been developed directly or indirectly as extensions of the classical Mohr–Coulomb criterion. Among them, the Mogi Coulomb criterion \cite{al2005relation}  developed from the linear Mogi and Coulomb criteria is one of the most widely applied. It express failure in terms of octahedral shear and mean normal stress to capture ($\sigma_2$) influence effectively under true triaxial condition. 
\par
In 2023, a new extended Mohr- Coulomb failure criterion \cite{mahetaji2023new} was developed by considering the weighting average of principal stresses for mean effective stress instead of the average of principal stresses. And they claimed that the new extended criterion shows equivalent results to Mogi-Coulomb criterion. They introduced the effect of $\sigma_2$ by considering a weighting parameter P of it to the mean stress, assuming weighting of $\sigma_1$ and $\sigma_3$ equal to unity. However, this unity weighting assumption is not universally supported by experimental evidence or theoretical development. Criteria like \cite{paul1968generalized, meyer2013linear,zeng2018paul} represents failure as a linear function of all three principal stresses with three independent, identifiable and unequal coefficients of principal stresses. Experimental and theoretical evidence demonstrates that the three principal stresses do not necessarily contribute identically to rock failure \cite{mogi1971fracture, colmenares2002statistical}.
\par
In this study, we developed a generalised 3D criterion by considering the effective normal stress as the convex combinations of principal stresses along with their unequal constrained weighting coefficients ($W_1\ne W_2\ne W_3$), subject to $W_i\geq0$, and $\sum W_i=1$. The new criteria considers the weighting of $\sigma_2$ on normal stress instead of mean stress. We choose available 11 rock types having both triaxial and true-triaxial data sets from the literatures for the calibration of the new criterion in principal stress ($\sigma_1, \sigma_2, \sigma_3$) space. The predicted results from the new criteria are compared and validated with the widely used criteria like Drucker-Prager, Mogi-Coulomb and Extended Mohr-Coulomb, in Haigh westergaard coordinates also. 

\section{Haigh-Westergaard Convention}
\begin{figure}[htbp]
    \centering
    \includegraphics[width=\textwidth]{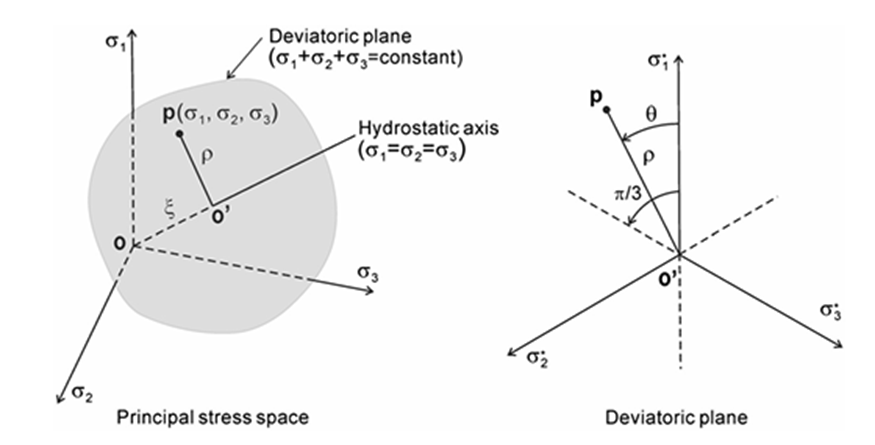}
    \caption{Geometric representation of a stress state in principal stress space and Haigh-Westergaard space.\cite{lee2012failure}}
    \label{fig:Haigh_Wester}
\end{figure}

The Haigh–Westergaard stress space provides a convenient representation and visualization of the three-dimensional stress state, where the principal stress coordinates ($\sigma_1,\sigma_2,\sigma_3$) are transformed into the ($\xi,\rho,\theta$) coordinates \cite{chakrabarty2012theory,ottosen1977failure}. This representation separates the stress state into its hydrostatic and deviatoric components and provides a useful framework for examining the impact of the intermediate principal stress and the Lode angle on rock failure. The Haigh–Westergaard coordinates define the principal stresses in $\sigma_1\ge\sigma_2\ge\sigma_3$ order and geometrically their relationship are shown in Figure(\ref{fig:Haigh_Wester}). The coordinate $\xi$ represents the projection of the stress point onto the hydrostatic axis, whose unit vector is $\bar e=(1,1,1)/\sqrt{3}$, whereas $\rho$ represents the radial distance of the failure point from the hydrostatic axis on the deviatoric plane. The third coordinate, $\theta$, is the Lode angle, which defines the angular position of the stress state on the deviatoric plane, orthogonal to the hydrostatic axis. In terms of the principal stresses, these coordinates may be expressed as:
\begin{equation}
    \xi=\frac{I_1}{\sqrt{3}}=\frac{\sigma_1+\sigma_2+\sigma_3}{\sqrt{3}}
\end{equation}
\begin{equation}
    \rho=\sqrt{2J_2}=\frac{\sqrt{(\sigma_1-\sigma_2)^2+(\sigma_2-\sigma_3)^2+(\sigma_3-\sigma_1)^2}}{\sqrt{3}}
\end{equation}
\begin{equation}
    \theta=\cos^{-1}\left(\frac{2\sigma_1-\sigma_2-\sigma_3}{2\sqrt{3}\sqrt{J_2}}\right)
\end{equation}
\begin{equation}
    J_2=\frac{1}{6}[(\sigma_1-\sigma_2)^2+(\sigma_2-\sigma_3)^2+(\sigma_3-\sigma_1)^2]
\end{equation}
For a fixed value of $\theta$ we get 2D plane ($\xi,\rho$) called Rendulic plane. At fixed value of $\xi$ we get $\pi$ plane.The principal stresses ($\sigma_1,\sigma_2,\sigma_3$) and ($\xi,\rho,\theta$) are related by the following relations \cite{brannon2009kayenta}.
\[
\begin{Bmatrix}
\sigma_1\\[2mm]
\sigma_2\\[2mm]
\sigma_3
\end{Bmatrix}
=
\begin{Bmatrix}
\dfrac{\xi}{\sqrt{3}}\\[2mm]
\dfrac{\xi}{\sqrt{3}}\\[2mm]
\dfrac{\xi}{\sqrt{3}}
\end{Bmatrix}
+
\sqrt{\frac{2}{3}}\rho
\begin{Bmatrix}
\frac{\sqrt{3}\cos\theta-\sin\theta}{2}\\[2mm]
\sin\theta\\[2mm]
\frac{-\sqrt{3}\cos\theta-\sin\theta}{2}
\end{Bmatrix}
\]
The stress angle $\theta$ ranges from $0$ to $\frac{\pi}{3}$ and $\cos(3\theta)$ is also called the Lode parameter.
\par
\section{Formulation of the New Generalised Three-Dimensional Criterion}
For rock materials, we generally deal with the principal stresses in the order $\sigma_1\ge\sigma_2\ge\sigma_3$ and treat compressive stress as positive and tensile stress as negative \cite{jiang2018simple}. The Mohr-Moulomb criteria with two parameters provide a linear relationship between shear strength of material and normal stress which can be express as given by \cite{labuz2014mohr} 
\begin{equation}
    \tau=\sigma\, \tan\phi + c
    \label{eqn1}
\end{equation}
Here $\phi$ is the internal friction angle, $c$ is the cohesion of material, $\tau$ is the shear strength and $\sigma$ is the applied normal stress.
It can be expressed in terms of principal stresses as: 
\begin{equation}
    \sigma_1=\frac{1+\sin\phi}{1-\sin\phi}\sigma_3 +\frac{2c\cos\phi}{1-\sin\phi}
    \label{eqn2}
\end{equation}
or \begin{equation}
    \sigma_1= q\cdot\sigma_3 + \text{UCS}
\end{equation}
From Mohr circle we know,
   $$ \sigma_n= \frac{\sigma_1+\sigma_3}{2}+\frac{\sigma_1-\sigma_3}{2}\cos2\alpha  \quad
    \text{and} \quad
\tau=\frac{\sigma_1-\sigma_3}{2}\sin 2\alpha$$
here $\alpha$ is angle of failure and $2\alpha=\phi+90^{\circ}$, 
$$ \sigma_n=\sigma_m-\tau_m \sin\phi \quad \text{and} \quad \tau=\tau_m\cos\phi $$
$\sigma_m=\frac{\sigma_1+\sigma_3}{2}$ is the mean stress and $\tau_m=\frac{\sigma_1-\sigma_3}{2}$ is the maximum stress. The normal stress can be written as
\begin{equation}
     \sigma_n= \frac{\sigma_1+\sigma_3}{2}-\frac{\sigma_1-\sigma_3}{2}\sin\phi
\end{equation}
\begin{equation}
    \sigma_n=\frac{(1-\sin\phi)}{2}\sigma_1+\frac{(1+\sin\phi)}{2}\sigma_3
\end{equation}
\begin{equation}
    \sigma_n=\frac{\sigma_1\left(\frac{1-\sin\phi}{2}\right)+\sigma_3\left(\frac{1+\sin\phi}{2}\right)}{\frac{1-\sin\phi}{2}+\frac{1+\sin\phi}{2}}
\end{equation}
So, for two principal stresses $\sigma_1$ and $\sigma_3$ acting on mutually perpendicular  palnes the normal stress can be written as, 
\begin{equation}
    \sigma_n=\frac{\sigma_1W_1+\sigma_3W_3}{W_1+W_3}
\end{equation}
\par
Where $W_1=\frac{1-\sin\phi}{2}$ and $W_3=\frac{1+\sin\phi}{2}$ can be consider as the weighting of the principal stresses $\sigma_1$ and $\sigma_3$ respectively ($W_1+W_3=1$). Similarly, the normal stress acting on an arbitrarily oriented plane in a three-dimensional principal stress state can be expressed as convex combinations of principal stresses \cite{jaeger1969fundamentals}.
\begin{equation}
    \sigma_n=\frac{\sigma_1W_1+\sigma_2 W_2+\sigma_3W_3}{W_1+W_2+W_3}
\end{equation}
In deriving our new generalised criterion we consider the impact of intermediate principal stress ($\sigma_2$) on rock strength as a weighting ($W_2=K$) of ($\sigma_2$). Also the contribution of $\sigma_1$ and $\sigma_3$ as a weighting $W_1$ and $W_3$ respectively. Where $W_2=K$, $W_1=(1-K)\frac{1-\sin\phi}{2}$ and $W_3=(1-K)\frac{1+\sin\phi}{2}$. K is the weighting of intermediate principal stress $\sigma_2$ and $\phi$ is the 3D frictional angle that resist the motion. Taking into account the weights, the effective normal stress can be expressed by the given equation.
\begin{equation}
    \sigma_n=\sigma_1W_1+\sigma_2W_2+\sigma_3W_3
\end{equation}
\begin{equation}
    \sigma_n=(1-K)\left[ \sigma_1\frac{(1-\sin\phi)}{2}+\sigma_3\frac{(1+\sin\phi)}{2}\right]+\sigma_2K
\end{equation}
Now considering the maximum shear stress and effective normal stress, the Mohr-Coulomb equation (\ref{eqn1}) reduce to 
\begin{equation}
    \tau_m\cos\phi=(1-K)\left[ \sigma_1\frac{(1-\sin\phi)}{2}+\sigma_3\frac{(1+\sin\phi)}{2}\right]\tan\phi+\sigma_2K\tan\phi+c
\end{equation}
i.e.,
\begin{equation}
    \frac{\sigma_1-\sigma_3}{2}\cos\phi=(1-K)\left[ \sigma_1\frac{(1-\sin\phi)}{2}+\sigma_3\frac{(1+\sin\phi)}{2}\right]\tan\phi+\sigma_2K\tan\phi+c
\end{equation}
By simplifying the above equation we can get the New Generalised 3D criterion with substancially different contributions of principal stresses via weights at failure. 
\begin{equation}
 \begin{aligned}   
    \sigma_1=\frac{2K\sin\phi}{(1-\sin\phi)(1+k\sin\phi)}\sigma_2+\frac{(1+\sin\phi)(1-k\sin\phi)}{(1-\sin\phi)(1+k\sin\phi)}\sigma_3\\  +\frac{2c\cos\phi}{(1-\sin\phi)(1+k\sin\phi)}
 \end{aligned}  
 \label{eqn17}
\end{equation}
Or, 
\begin{equation}
    \sigma_1=A\sigma_2+B\sigma_3+CS
    \label{new-eq}
\end{equation}
Here,
\begin{align*}
A &= \frac{2K\sin\phi}
{(1-\sin\phi)(1+k\sin\phi)},\\
B &= \frac{(1+\sin\phi)(1-k\sin\phi)}
{(1-\sin\phi)(1+k\sin\phi)},\\
CS &= \frac{2c\cos\phi}
{(1-\sin\phi)(1+k\sin\phi)}.
\end{align*}
This newly developed eqns(13,14) are derived from Mohr-Coulomb by considering the effective normal stress in 3D as the convex combinations of principal stresses with their weights having unequal influences at failure. $K,c$ and $\phi$ are the parameters of the new equation. $K$ is the coefficient of the intermediate principal stress and is a function of $c$ and $\phi$. $c$ is defined as the cohesion or inherent strength of rock and $\phi$ is the 3D frictional angle that resist sliding.
\par
If we neglect the effect of the intermediate principal stress (i.e. $K=0$), the new equaion will have $A=0$, $B=\frac{1+\sin\phi}{1-\sin\phi}$ and $CS=\frac{2c\cos\phi}{1-\sin\phi}$ which is exactly the Mohr-Coulomb failure criteria as in eqn(\ref{eqn2}).
\begin{equation*}
    \sigma_1 =
    \frac{1+\sin\phi}{1-\sin\phi}\sigma_3
    + \frac{2c\cos\phi}{1-\sin\phi}
\end{equation*}
Therefore, the New Generalised criterion coincides with the Mohr-Coulomb failure criteria if the effect of the intermediate principal stress is ignored. The parameter K determines the proportionate contribution of the intermediate principal stress to the failure situation. Its variation affects the coefficients related to ($\sigma_2$), ($\sigma_3$), and the strength intercept ($CS$), altering the projected failure envelope under true-triaxial stress. The value of $K$ can be determined with the help of Mohr-Coulomb failure criteria to find $UCS$ as given 
\begin{equation}
    K=\frac{2c\cos\phi-UCS(1-\sin\phi)}{UCS\sin\phi(1-\sin\phi)}
\end{equation}
\par
The New Generalised criterion can also be expressed in Haigh-Westergaard coordinates as 
\begin{equation}
   \rho= \frac{\sqrt{2}(A+B-1)\xi+\sqrt{6}CS}{\sqrt{3}(1+B)\cos\theta+(B-2A-1)\sin\theta}
\label{eqn20}
\end{equation}
\par
The compressive meridian ($\theta=0$) and the extension meridian ($\theta=\frac{\pi}{3}$) are given by eqns~(\ref{eqn21}) and~(\ref{eqn22}). The parameter $K$ governs the magnitude and meridional shape of the failure surface. Increasing $K$ increases both the meridians with a stronger effect on the extension meridian.
\begin{equation}
    \rho_c=\frac{(\sqrt{2}\sin\phi)\xi+\sqrt{6}c\cos\phi}{\sqrt{3}(1-K\sin^2\phi)}
    \label{eqn21}
\end{equation}
\begin{equation}
    \rho_t=\frac{2[(\sqrt{2}\sin\phi)\xi+\sqrt{6}c\cos\phi]}{\sqrt{3}[(1+\sin\phi)-K(3\sin\phi+\sin^2\phi)]}
    \label{eqn22}
\end{equation}
\section{Calibration of the proposed failure criterion}
 For intact rocks, compressive failure and fracture are governed by the combined effects of compression and tension. In this study, experimental data on rock compressive strength obtained under both true-triaxial and triaxial stress states are considered. The triaxial compression test data were used to calibrate the proposed failure criterion. The best values of parameters $K$ and $\phi$ were obtained through linear regression by minimizing the misfit between the predicted and experimental values.

Eleven rock types reported in the literature are selected for the calibration and validation of the proposed K-model, as presented in Table(\ref{tab:rock_sources}).

\begin{table}[htbp]
\centering
\small
\setlength{\tabcolsep}{4pt}
\renewcommand{\arraystretch}{1.15}
\caption{Rock types and sources of experimental triaxial compression data used for calibration and validation.}
\label{tab:rock_sources}
\begin{tabularx}{\linewidth}{c l >{\raggedright\arraybackslash}X}
\toprule
\textbf{S. No.} & \textbf{Rock type} & \textbf{Experimental data source} \\
\midrule
1  & KTB amphibolite     & Colmenares and Zoback (2002)\cite{colmenares2002statistical}, presented and provided by Haimson and Chang (2000)\cite{haimson2000new}. \\
2  & Dunham dolomite     & Mogi (2006)\cite{mogi2006experimental}. \\
3  & Solenhofen limestone & Mogi (2006)\cite{mogi2006experimental}. \\
4  & Shirahama sandstone & Colmenares and Zoback (2002)\cite{colmenares2002statistical}, digitized from Takahashi and Koide (1989)\cite{takahashi1989effect}. \\
5  & Yuubari shale       & Colmenares and Zoback (2002)\cite{colmenares2002statistical}, digitized from Takahashi and Koide (1989)\cite{takahashi1989effect}. \\
6  & Yamaguchi marble    & Mogi (2006)\cite{mogi2006experimental}. \\
7  & Mizuho trachyte     & Mogi (2006)\cite{mogi2006experimental}. \\
8  & Manazuru andesite   & Mogi (2006)\cite{mogi2006experimental}. \\
9  & Inada granite       & Mogi (2006)\cite{mogi2006experimental}. \\
10 & Orikabe monzonite   & Mogi (2006)\cite{mogi2006experimental}. \\
11 & Coconino sandstone  & Xiaodong and Haimson (2016)\cite{ma2016failure}. \\
\bottomrule
\end{tabularx}
\end{table}
\par
The predictive accuracy of the proposed model is evaluated with the help of Root Mean Square Error(RMSE). RMSE is defined as the square root of the mean of the squared error between experimental stress and predicted strength by model.
\begin{table}[htbp]
\centering
\caption{Calibration result using the triaxial compression test for eleven rock types}
\label{tab:calibration}

\small
\renewcommand{\arraystretch}{1.15}
\setlength{\tabcolsep}{4pt}

\begin{tabularx}{\linewidth}{@{}lccccc@{}}
\toprule
Rock type
& $K$
& $\phi$
& $c$ (MPa)
& $UCS$
& \shortstack[c]{RMSE of $\sigma_1$\\by proposed criterion\\(MPa)} \\
\midrule
KTB amphibolite      & 0.11  & $46^\circ$ & 62.51  & 289.66 & 54.61 \\
Dunham dolomite      & 0.22  & $34^\circ$ & 113.90 & 379.30 & 33.95 \\
Solenhofen limestone  & 0.19  & $30^\circ$ & 107.36 & 339.12 & 22.34 \\
Shirahama sandstone  & 0.07  & $38^\circ$ & 21.82  & 86.18  & 7.41  \\
Yuubari shale        & 0.13  & $30^\circ$ & 31.92  & 103.18 & 5.39  \\
Yamaguchi marble     & 0.14  & $38^\circ$ & 30.10  & 114.01 & 17.87 \\
Mizuho trachyte      & 0.11  & $31^\circ$ & 49.41  & 165.30 & 15.38 \\
Manazuru andesite    & 0.17  & $44^\circ$ & 71.16  & 301.75 & 35.87 \\
Inada granite        & 0.26  & $44^\circ$ & 116.57 & 461.50 & 44.72 \\
Orikabe monzonite    & 0.38  & $38^\circ$ & 139.41 & 461.17 & 48.98 \\
Coconino sandstone   & 0.03  & $32^\circ$ & 45.47  & 161.05 & 26.97 \\
\bottomrule
\end{tabularx}

\end{table}

\par
\begin{equation}
    RMSE=\sqrt{\frac{1}{n}\sum_{i}^{n}e_i^2} \quad i=1,2,...,n
\end{equation}
\par
Where n is the number of experimental data points of a particular rock. $e_i$ is the error defined as 
\begin{equation}
    e_i=\sigma_i^{pre}-\sigma_i^{exp}
\end{equation}
$\sigma_i^{pre}$ is $i^{th}$ predicted value by the proposed model and $\sigma_i^{exp}$ is the $i^{th}$ experimental data point. A smaller RMSE denotes a better agreement between predicted and experimental data.
\par
\begin{figure}[tbp]
    \centering
    \includegraphics[width=\linewidth]{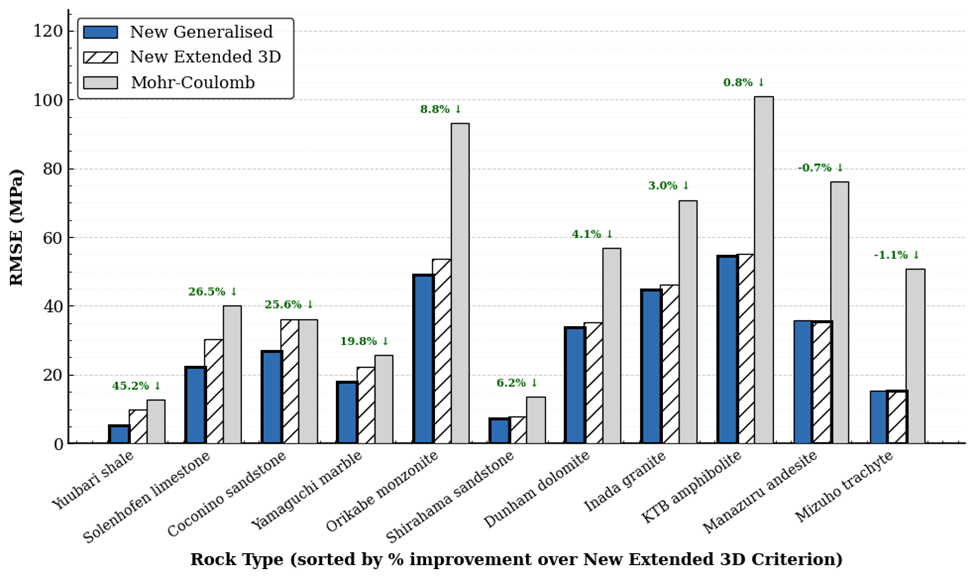}
    \caption{Comparison of RMSE for $\sigma_1$ in triaxial compression test data.}
    \label{fig:rmse-sigma1}
\end{figure}
The triaxial test data for eleven rock types were fitted to the proposed new Generalised 3D failure criterion. The model coefficients A, B and $CS$ were calculated using a least-squares optimization approach by minimizing the error between the experimental and predicted strength.
\par
The calibration results on rock strength predicted by the new criterion on eleven rock types under triaxial compression are presented in figure(\ref{fig:calibration-triaxial}). The scatter points represent the experimental triaxial compression test data, while the solid lines illustrate the corresponding predictions obtained using eq(\ref{new-eq}).

\begin{figure}[p]
    \centering

    \begin{subfigure}[t]{0.48\textwidth}
        \centering
        \includegraphics[width=\linewidth,height=0.26\textheight,keepaspectratio]{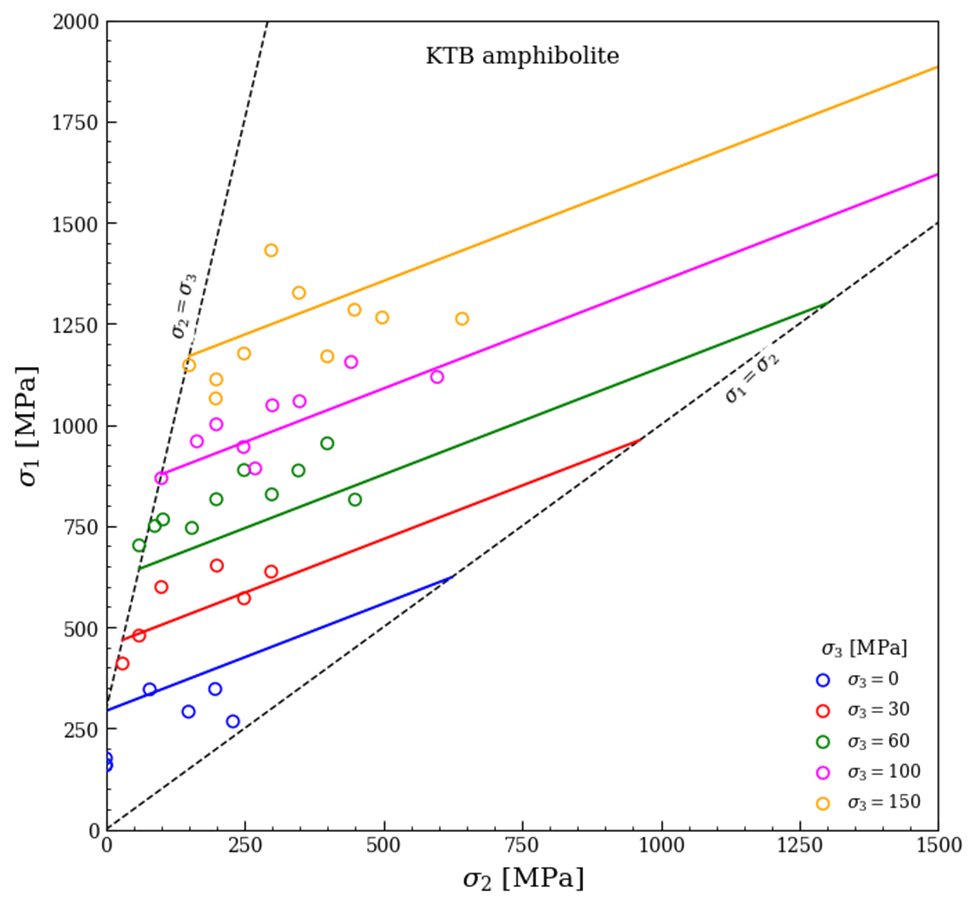}
        \caption{}
        \label{fig:calibration-a}
    \end{subfigure}
    \hfill
    \begin{subfigure}[t]{0.48\textwidth}
        \centering
        \includegraphics[width=\linewidth,height=0.26\textheight,keepaspectratio]{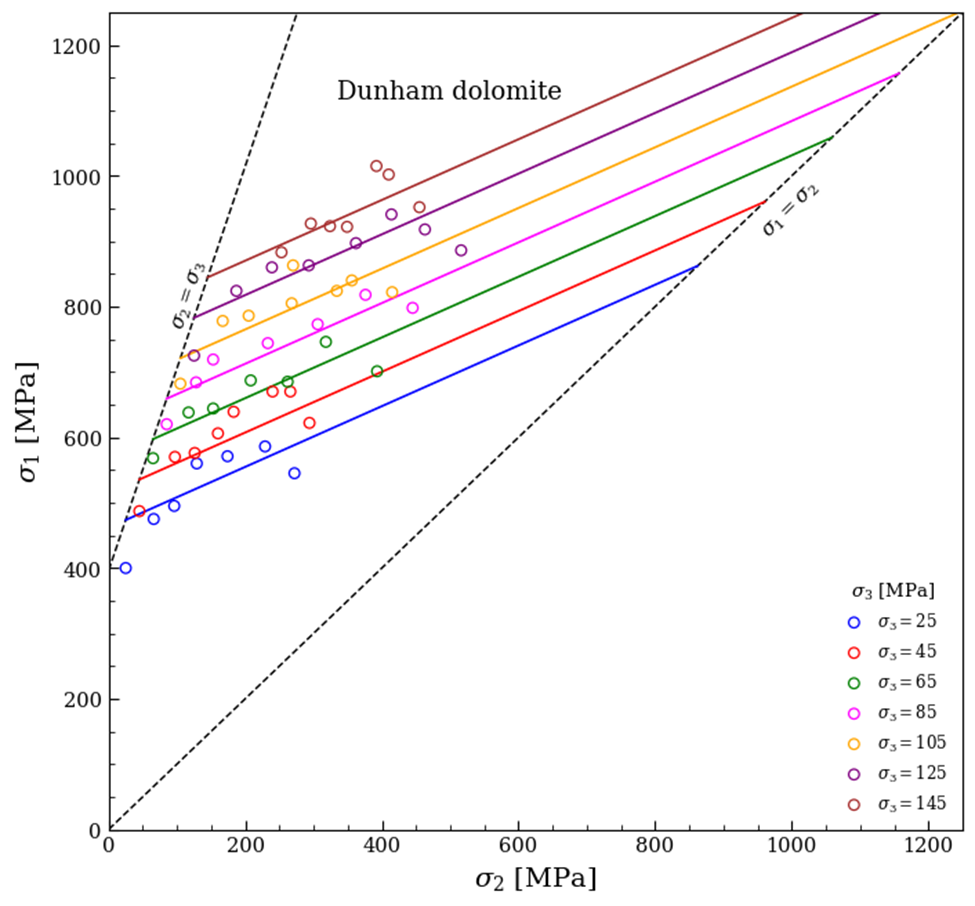}
        \caption{}
        \label{fig:calibration-b}
    \end{subfigure}

    \medskip

    \begin{subfigure}[t]{0.48\textwidth}
        \centering
        \includegraphics[width=\linewidth,height=0.26\textheight,keepaspectratio]{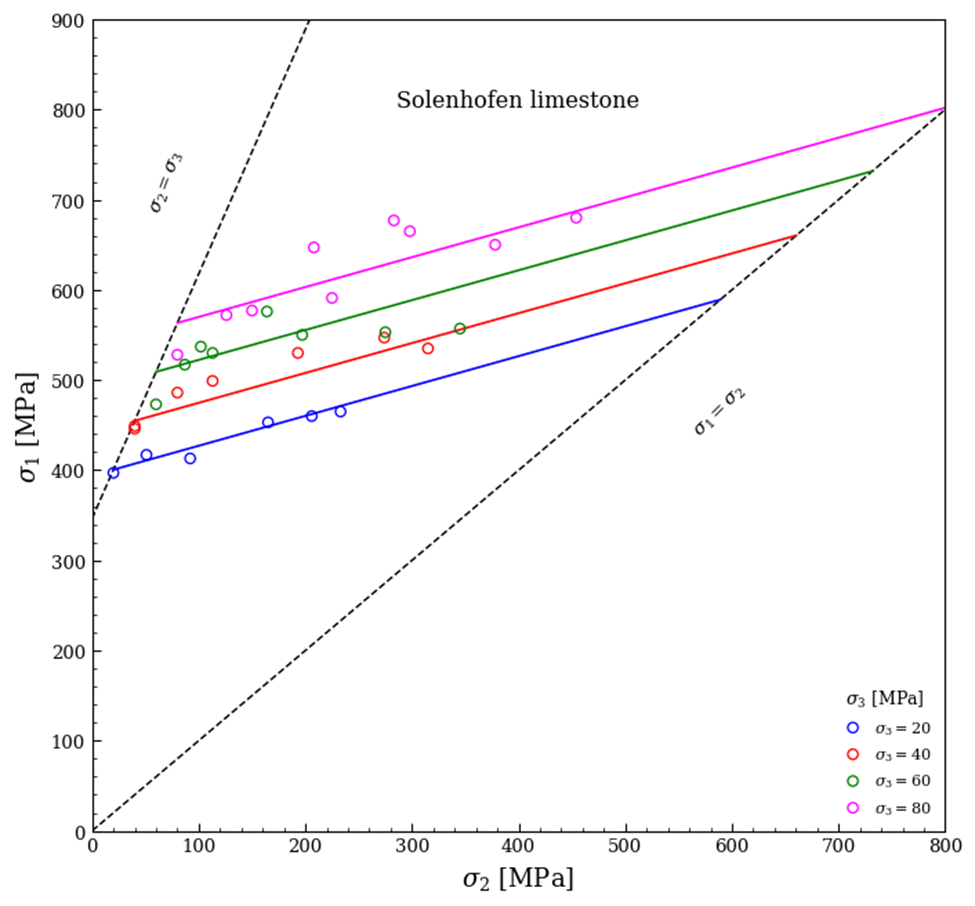}
        \caption{}
        \label{fig:calibration-c}
    \end{subfigure}
    \hfill
    \begin{subfigure}[t]{0.48\textwidth}
        \centering
        \includegraphics[width=\linewidth,height=0.26\textheight,keepaspectratio]{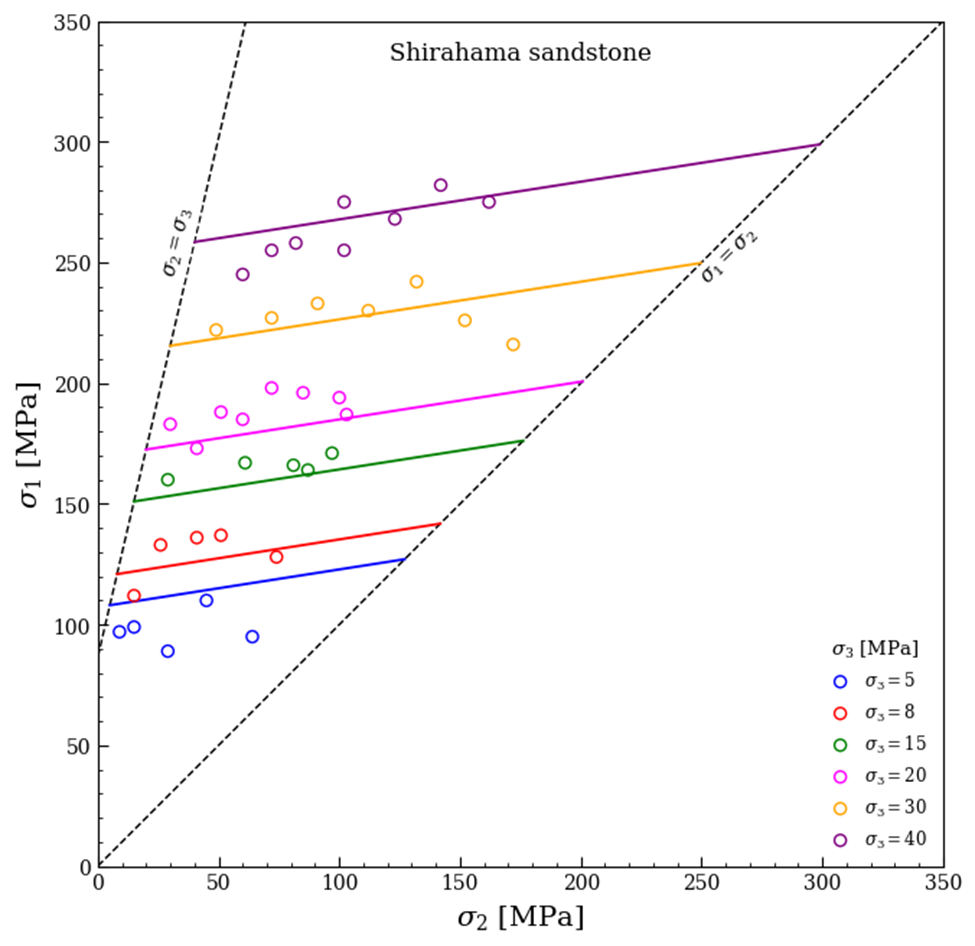}
        \caption{}
        \label{fig:calibration-d}
    \end{subfigure}

    \medskip

    \begin{subfigure}[t]{0.48\textwidth}
        \centering
        \includegraphics[width=\linewidth,height=0.26\textheight,keepaspectratio]{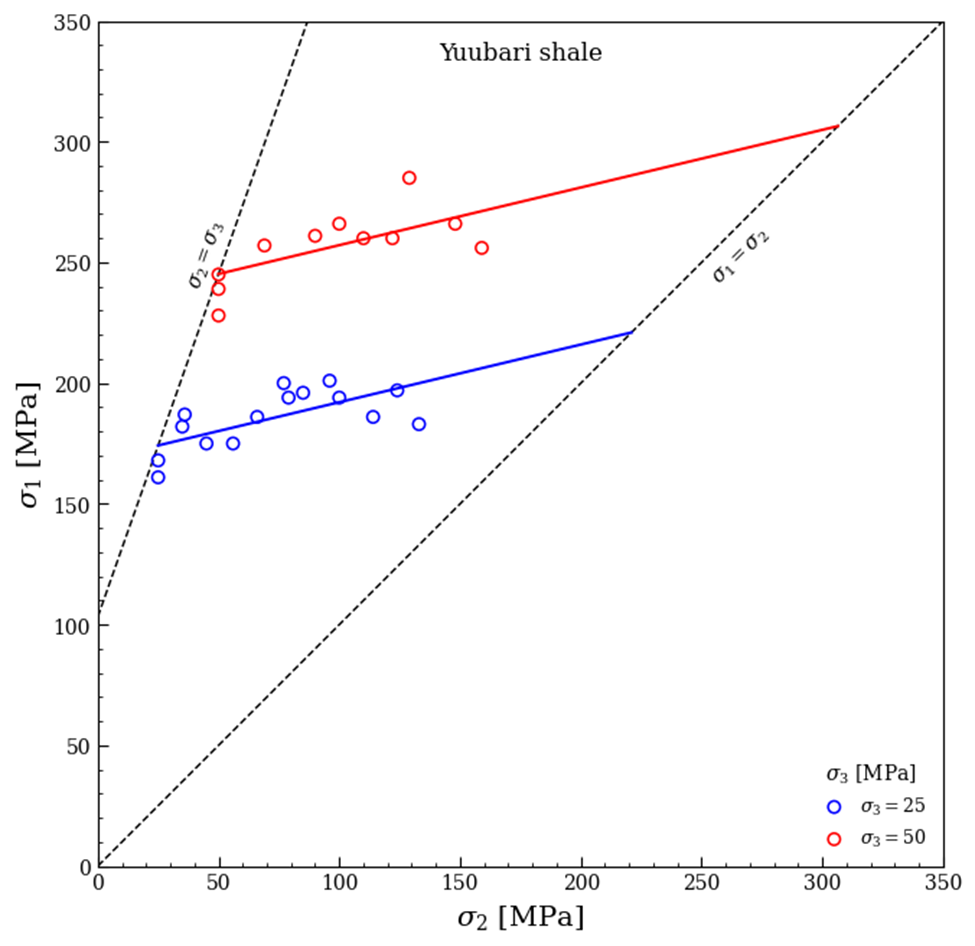}
        \caption{}
        \label{fig:calibration-e}
    \end{subfigure}
    \hfill
    \begin{subfigure}[t]{0.48\textwidth}
        \centering
        \includegraphics[width=\linewidth,height=0.26\textheight,keepaspectratio]{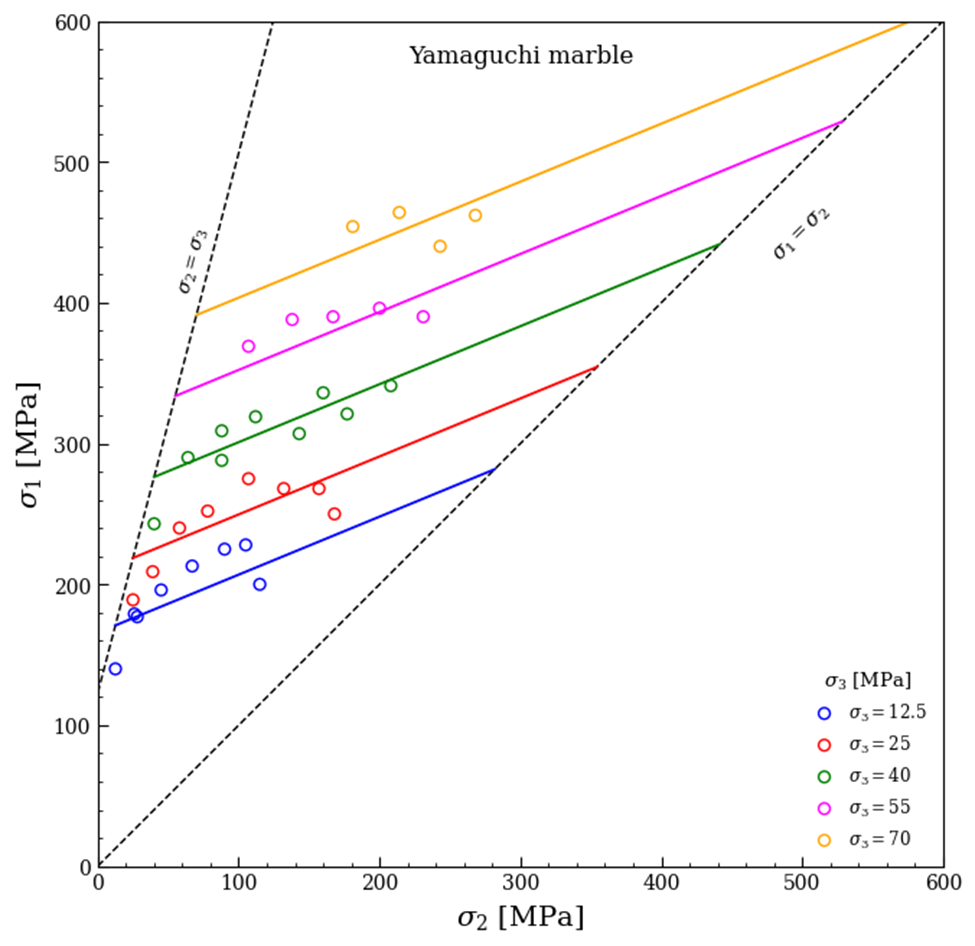}
        \caption{}
        \label{fig:calibration-f}
    \end{subfigure}

    \caption{Calibration of the proposed new criterion using triaxial compression data in the $\sigma_2-\sigma_1$ space for eleven rock types.}
    \label{fig:calibration-triaxial}
\end{figure}

\begin{figure}[p]
    \ContinuedFloat
    \centering

    \begin{subfigure}[t]{0.48\textwidth}
        \centering
        \includegraphics[width=\linewidth,height=0.26\textheight,keepaspectratio]{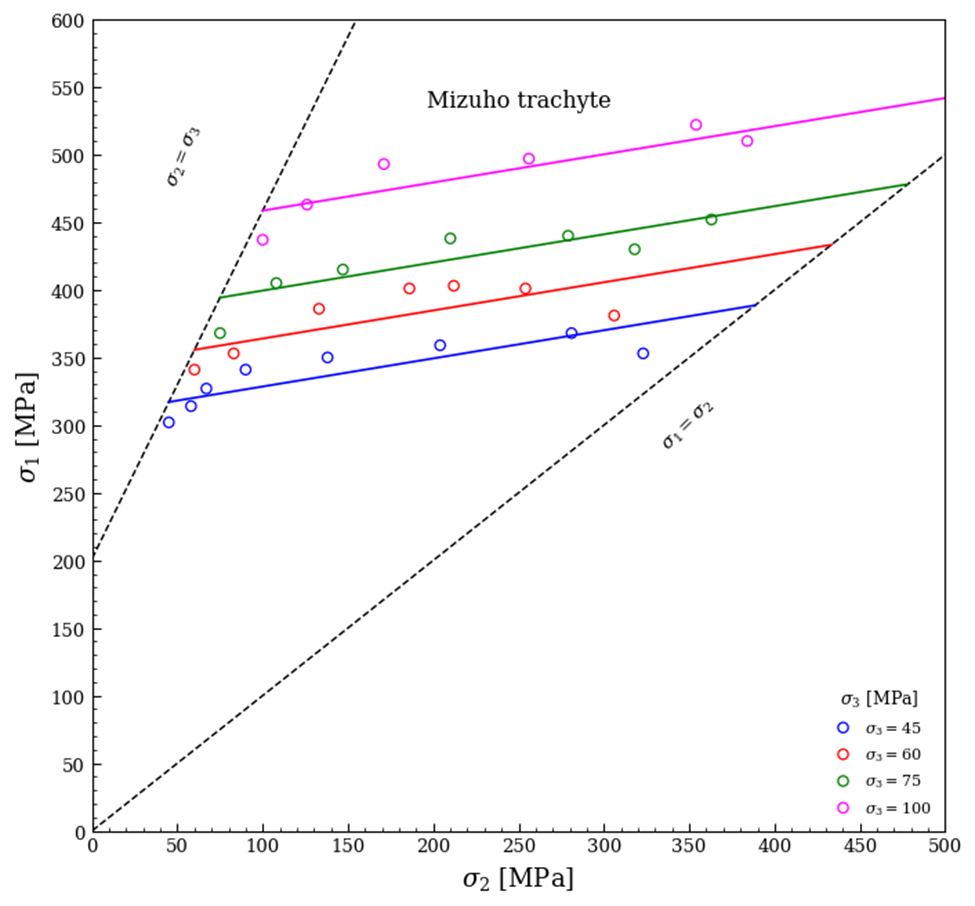}
        \caption{}
        \label{fig:calibration-g}
    \end{subfigure}
    \hfill
    \begin{subfigure}[t]{0.48\textwidth}
        \centering
        \includegraphics[width=\linewidth,height=0.26\textheight,keepaspectratio]{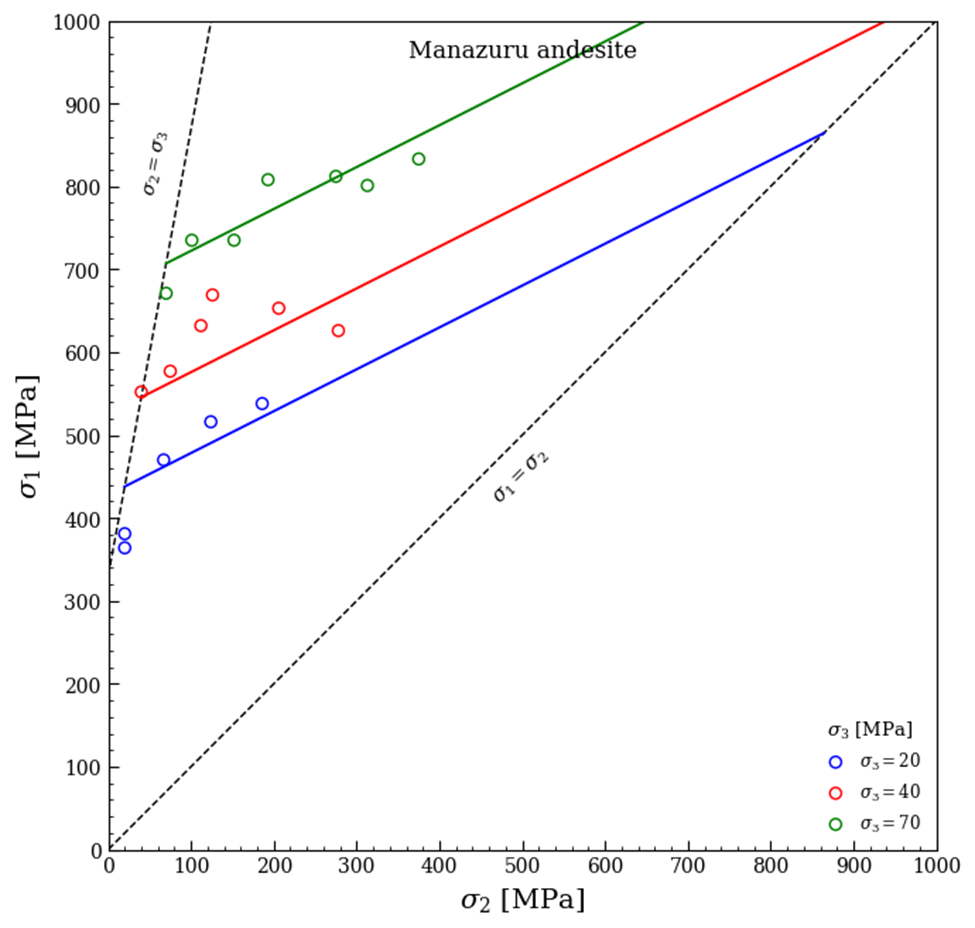}
        \caption{}
        \label{fig:calibration-h}
    \end{subfigure}

    \medskip

    \begin{subfigure}[t]{0.48\textwidth}
        \centering
        \includegraphics[width=\linewidth,height=0.26\textheight,keepaspectratio]{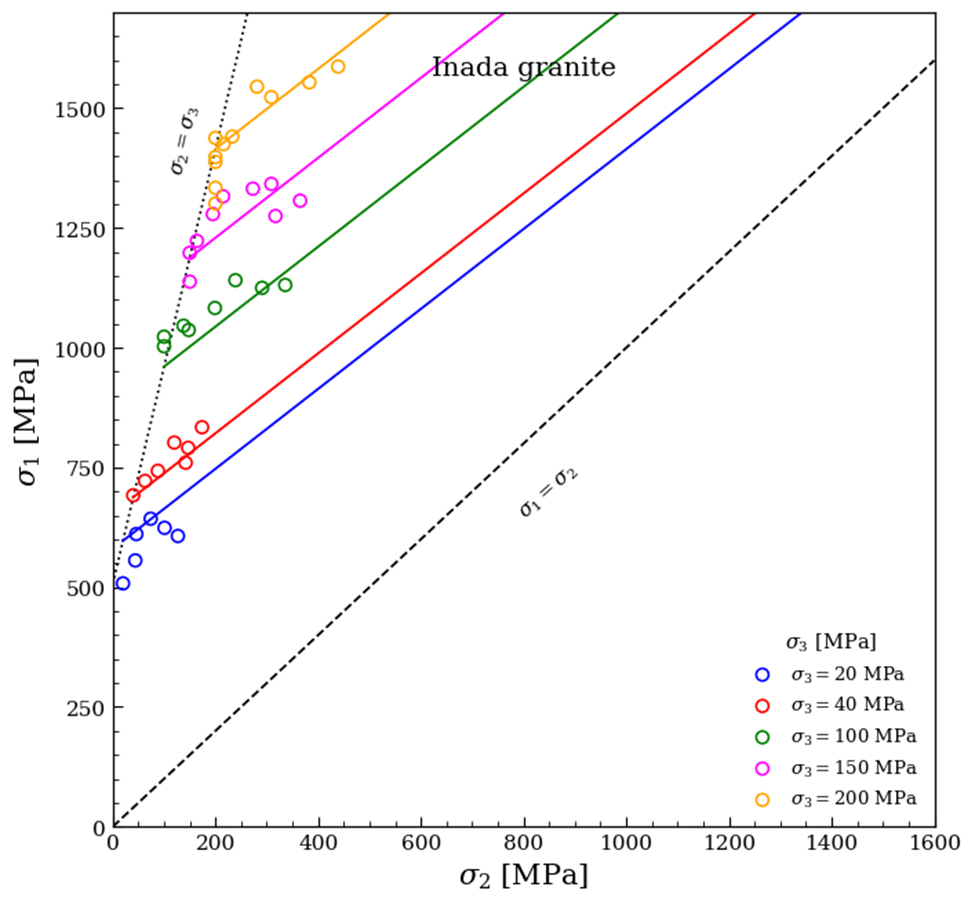}
        \caption{}
        \label{fig:calibration-i}
    \end{subfigure}
    \hfill
    \begin{subfigure}[t]{0.48\textwidth}
        \centering
        \includegraphics[width=\linewidth,height=0.26\textheight,keepaspectratio]{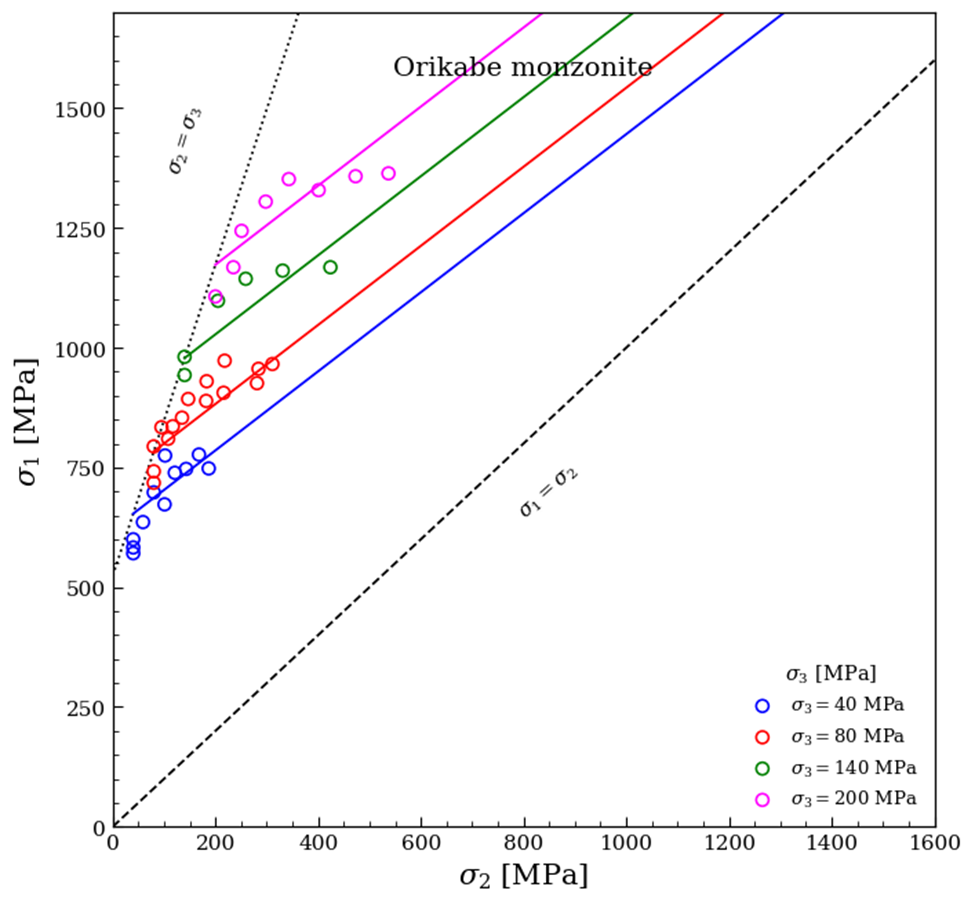}
        \caption{}
        \label{fig:calibration-j}
    \end{subfigure}

    \medskip

    \begin{subfigure}[t]{\textwidth}
        \centering
        \includegraphics[width=0.48\textwidth,height=0.26\textheight,keepaspectratio]{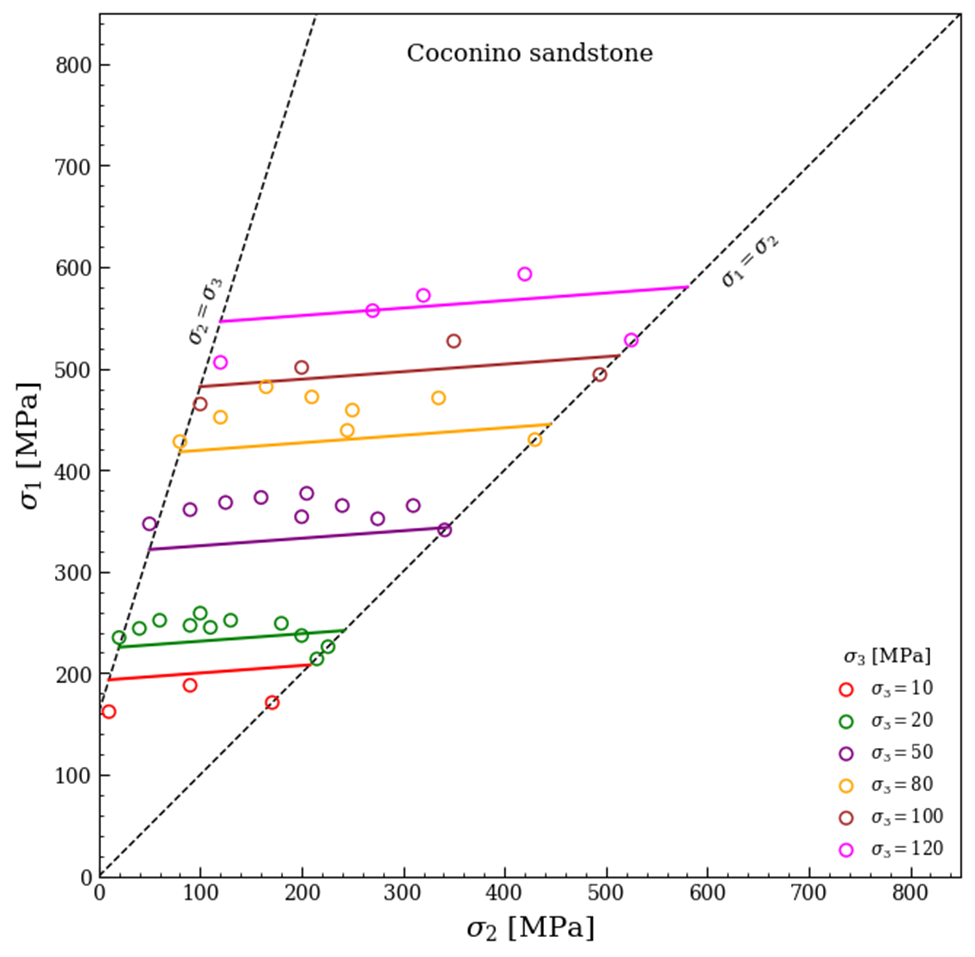}
        \caption{}
        \label{fig:calibration-k}
    \end{subfigure}

    \caption[]{Calibration of the proposed new criterion using triaxial compression data in the $\sigma_2-\sigma_1$ space for eleven rock types (continued).}
\end{figure}
\par
The calibration results shows that the influence of intermediate principal stress depends on rock type and stress conditions. At constant confining stress, the rock strength initially increases with increasing $\sigma_2$. However, at higher $\sigma_2$ the rock strength either decreases or reach a plateau.
\par
From the calibration, the values of the material parameters are determined and RMSE for each rock type are calculated. For the best fit result the RMSE for $\sigma_1$ must be minimum. The RMSE values of our proposed model for $\sigma_1$ is compared with Mohr-Coulomb and Extended Mohr-Coulomb as shown in fig(\ref{fig:rmse-sigma1}).The values of RMSE for $\sigma_1$ for our proposed model lies in between 5.39 MPa to 54.61 MPa range as given in table(\ref{tab:calibration}).
\section{Validation of the Proposed Failure Criterion in Haigh--Westergaard Space}In section-4, we have shown the calibration results and obtained the values of $A$, $B$ and CS with the help of eqn(\ref{new-eq}) along with linear regression of the triaxial test data in the plot of $\sigma_1$ vs $\sigma_2$. Later the values of $K$, $\phi$ and $C$(MPa) are calculated from eqn(\ref{eqn17}) and used for further validation in Haigh-Westergaard space.
\par
In $(\xi,\rho,\theta)$ space, $\rho$ is the dependent variable, $\xi$ is the independent variable and $\theta$ depends on rock type. The prediction is done with the help of eqn(\ref{eqn20}) and the result is compared with the experimental triaxial compression test data as shown in figure(\ref{fig:validation-xi-rho}), where solid line indicates the failure line by eqn(\ref{eqn20}) and scatter points are triaxial test data. For the validation of the proposed new criterion we use the Determination Coefficient (DC), RMSE and Error Percentage of $\rho$ for the predicted rock strength.
\subsection{Coefficient of Determination (DC).}
DC is a statistical scaler indicator that tells us about the reliability of the predicted data. DC of $\rho$ is defined as, 
\begin{equation}
    DC=1-\frac{\sum_{i}^n (\rho_i^{pred}-\rho_i^{test})^2 }{\sum_{i}^n (\rho_i^{test}-\bar\rho_{test})^2 }
\end{equation}
\par
Here, $\rho_i^{pred}$ are the $i^{th}$ predicted values by the model, $\rho_i^{test}$ are the $i^{th}$ experimental triaxial values and the $\bar\rho_{test}$ are the mean values of the experimental triaxial values. $DC<0$ indicates poorer agreement to experimental data and DC=0 is the baseline model.  A value of DC closer to 1 indicates stronger agreement with the experimental data and DC=1 indicates perfect agreement with zero misfit.

\begin{table*}[htbp]
\centering
\caption{Comparison of RMSE and DC values in terms of $\rho$ for the eleven rock types among NG--New Generalised criterion, NE--New Extended Mohr--Coulomb criterion, MG--Mogi--Coulomb criterion and DP--Drucker--Prager criterion.}
\label{tab:rmse_dc_rho}

\renewcommand{\arraystretch}{1.2}
\setlength{\tabcolsep}{4pt}
\scriptsize

\resizebox{\textwidth}{!}{%
\begin{tabular}{l c cccc cccc}
\toprule
\multirow{2}{*}{\textbf{Rock types}} &
\multirow{2}{*}{\textbf{No. of data}} &
\multicolumn{4}{c}{\textbf{RMSE ($\rho$)}} &
\multicolumn{4}{c}{\textbf{DC ($\rho$)}} \\
\cmidrule(lr){3-6} \cmidrule(lr){7-10}

& &
\textbf{NG} &
\textbf{NE} &
\textbf{MG} &
\textbf{DP} &
\textbf{NG} &
\textbf{NE} &
\textbf{MG} &
\textbf{DP} \\

\midrule
KTB amphibolite       & 42 & 36.62 & 39.96 & 40.64 & 73.63 & 0.97 & 0.98 & 0.97 & 0.89 \\
Dunham dolomite       & 54 & 22.70 & 24.34 & 16.22 & 47.88 & 0.92 & 0.95 & 0.96 & 0.71 \\
Solenhofen limestone  & 26 & 14.61 & 15.27 & 14.92 & 53.93 & 0.92 & 0.93 & 0.93 & 0.61 \\
Shirahama sandstone   & 38 & 9.30  & 10.76 & 11.06 & 14.89 & 0.92 & 0.91 & 0.87 & 0.78 \\
Yuubari shale         & 18 & 5.13  & 6.01  & 4.72  & 10.98 & 0.93 & 0.90 & 0.92 & 0.68 \\
Yamaguchi marble      & 38 & 12.49 & 14.50 & 8.36  & 18.86 & 0.96 & 0.92 & 0.95 & 0.90 \\
Mizuho trachyte       & 31 & 17.14 & 18.08 & 9.74  & 31.84 & 0.88 & 0.87 & 0.94 & 0.62 \\
Manazuru andesite     & 23 & 29.38 & 30.07 & 29.98 & 55.10 & 0.95 & 0.92 & 0.93 & 0.79 \\
Inada granite         & 46 & 30.78 & 34.59 & 36.16 & 61.04 & 0.98 & 0.98 & 0.97 & 0.92 \\
Orikabe monzonite     & 44 & 29.61 & 38.23 & 55.15 & 69.94 & 0.95 & 0.95 & 0.88 & 0.81 \\
Coconino sandstone    & 48 & 22.09 & 23.04 & 17.51 & 44.50 & 0.92 & 0.92 & 0.95 & 0.71 \\

\bottomrule
\end{tabular}%
}
\end{table*}

\begin{figure}[p]
    \centering

    \begin{subfigure}[t]{0.48\textwidth}
        \centering
        \includegraphics[width=\linewidth,height=0.26\textheight,keepaspectratio]{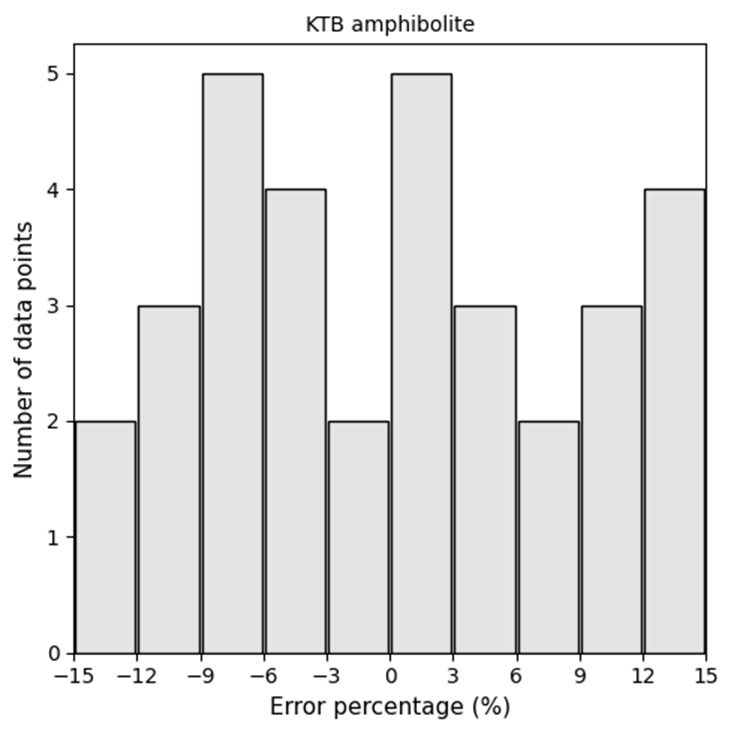}
        \caption{}
        \label{fig:validation-a}
    \end{subfigure}
    \hfill
    \begin{subfigure}[t]{0.48\textwidth}
        \centering
        \includegraphics[width=\linewidth,height=0.26\textheight,keepaspectratio]{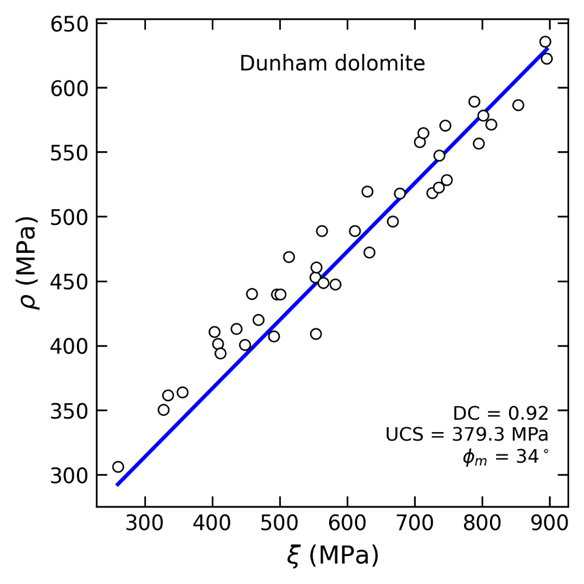}
        \caption{}
        \label{fig:validation-b}
    \end{subfigure}

    \medskip

    \begin{subfigure}[t]{0.48\textwidth}
        \centering
        \includegraphics[width=\linewidth,height=0.26\textheight,keepaspectratio]{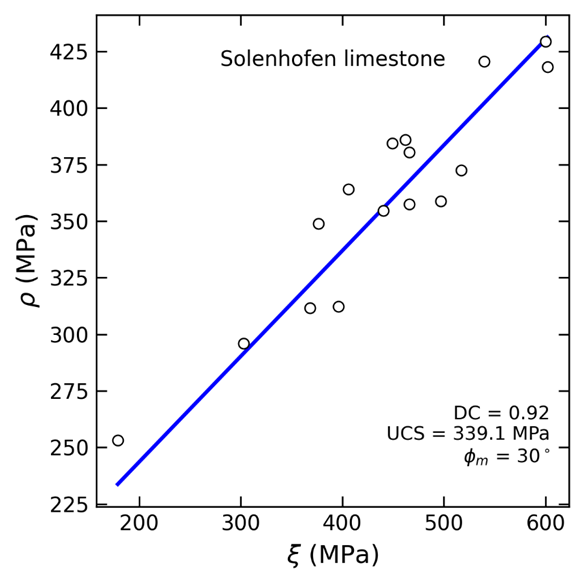}
        \caption{}
        \label{fig:validation-c}
    \end{subfigure}
    \hfill
    \begin{subfigure}[t]{0.48\textwidth}
        \centering
        \includegraphics[width=\linewidth,height=0.26\textheight,keepaspectratio]{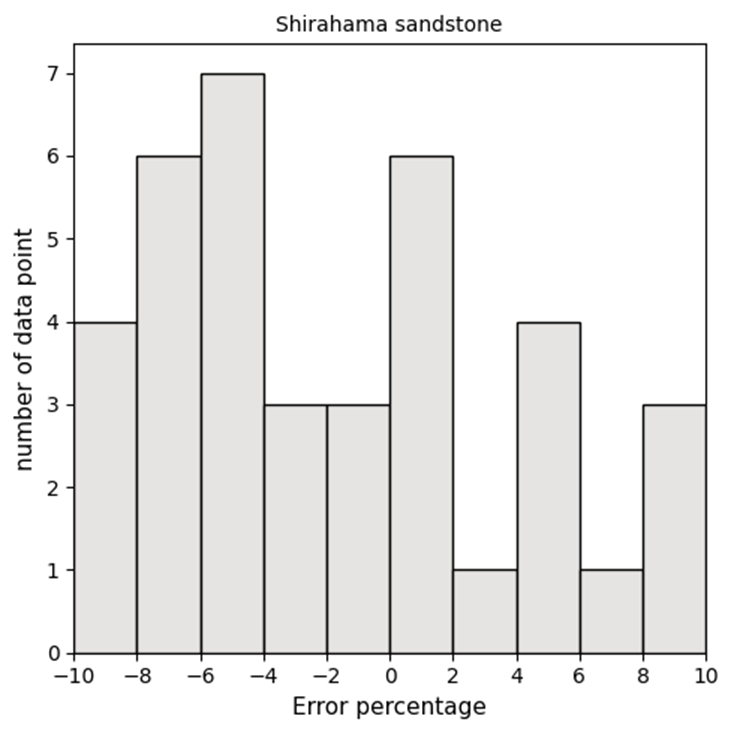}
        \caption{}
        \label{fig:validation-d}
    \end{subfigure}

    \medskip

    \begin{subfigure}[t]{0.48\textwidth}
        \centering
        \includegraphics[width=\linewidth,height=0.26\textheight,keepaspectratio]{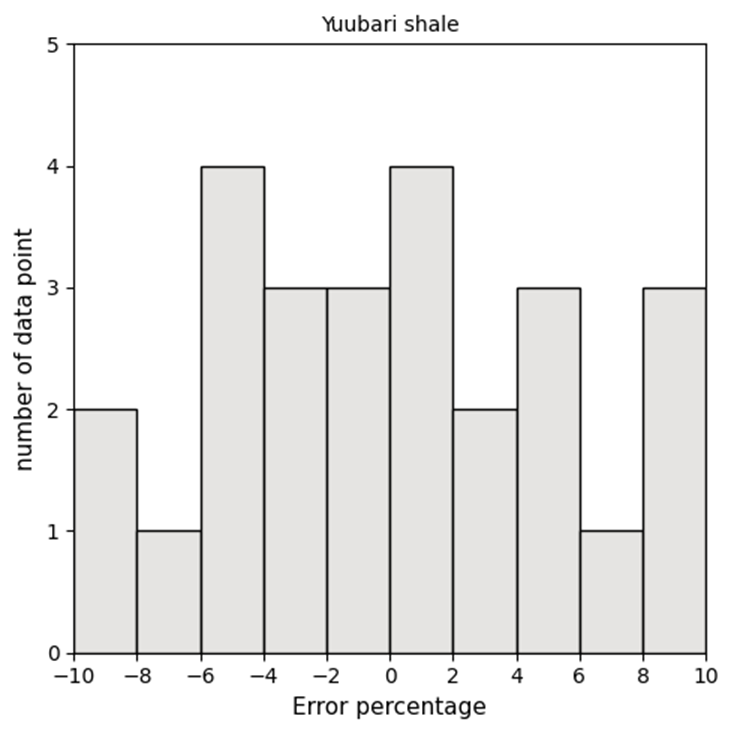}
        \caption{}
        \label{fig:validation-e}
    \end{subfigure}
    \hfill
    \begin{subfigure}[t]{0.48\textwidth}
        \centering
        \includegraphics[width=\linewidth,height=0.26\textheight,keepaspectratio]{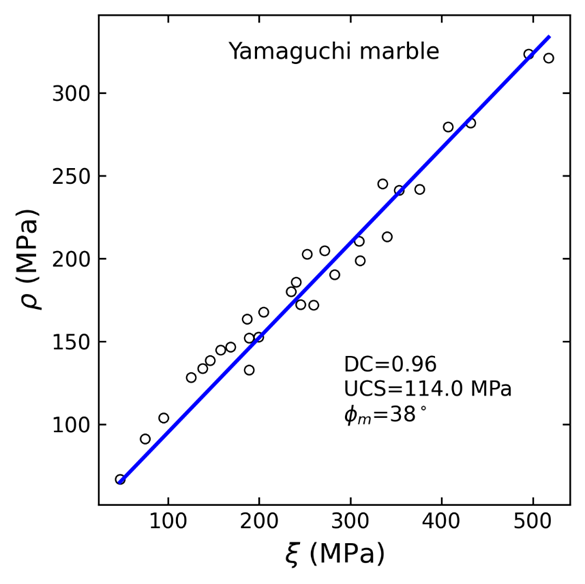}
        \caption{}
        \label{fig:validation-f}
    \end{subfigure}

    \caption{Validation plot in $\xi-\rho$ space for eleven rock types.}
    \label{fig:validation-xi-rho}
\end{figure}

\begin{figure}[p]
    \ContinuedFloat
    \centering

    \begin{subfigure}[t]{0.48\textwidth}
        \centering
        \includegraphics[width=\linewidth,height=0.26\textheight,keepaspectratio]{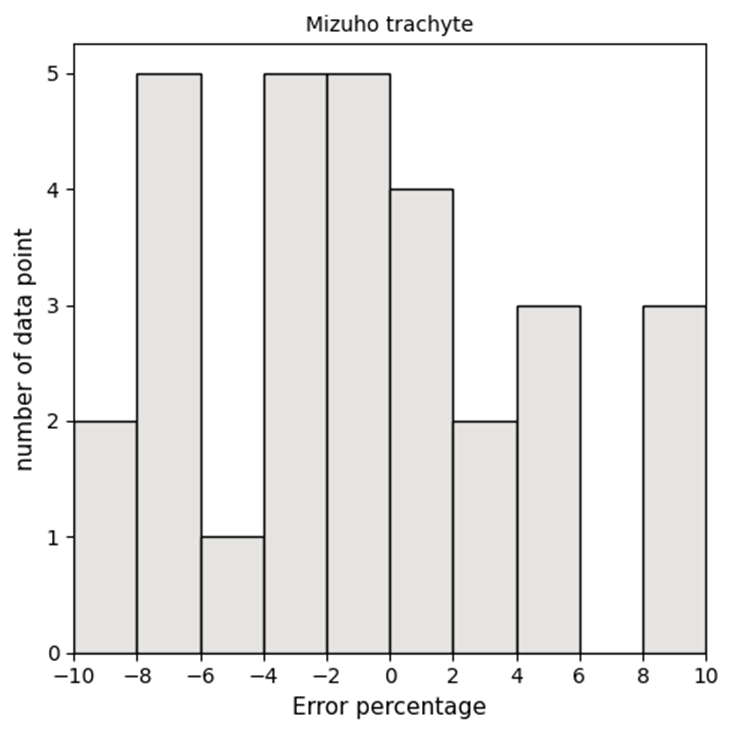}
        \caption{}
        \label{fig:validation-g}
    \end{subfigure}
    \hfill
    \begin{subfigure}[t]{0.48\textwidth}
        \centering
        \includegraphics[width=\linewidth,height=0.26\textheight,keepaspectratio]{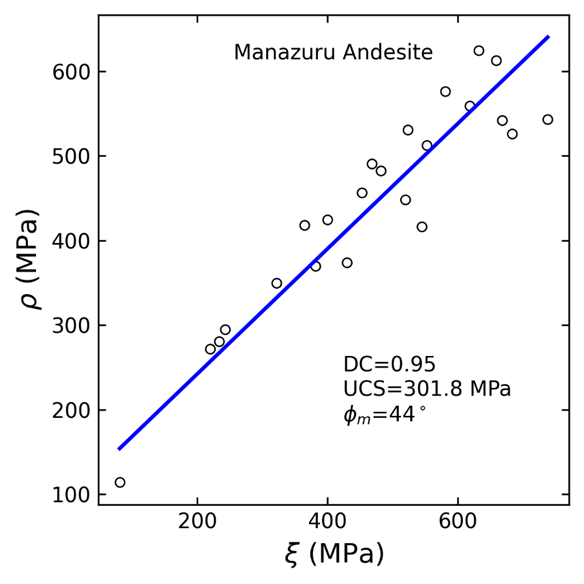}
        \caption{}
        \label{fig:validation-h}
    \end{subfigure}

    \medskip

    \begin{subfigure}[t]{0.48\textwidth}
        \centering
        \includegraphics[width=\linewidth,height=0.26\textheight,keepaspectratio]{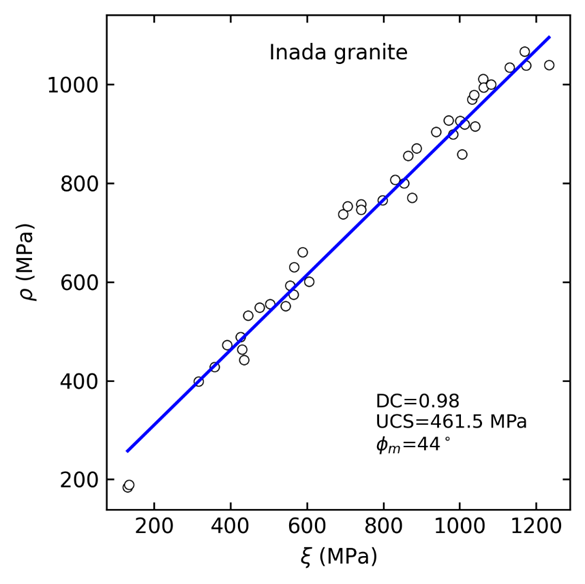}
        \caption{}
        \label{fig:validation-i}
    \end{subfigure}
    \hfill
    \begin{subfigure}[t]{0.48\textwidth}
        \centering
        \includegraphics[width=\linewidth,height=0.26\textheight,keepaspectratio]{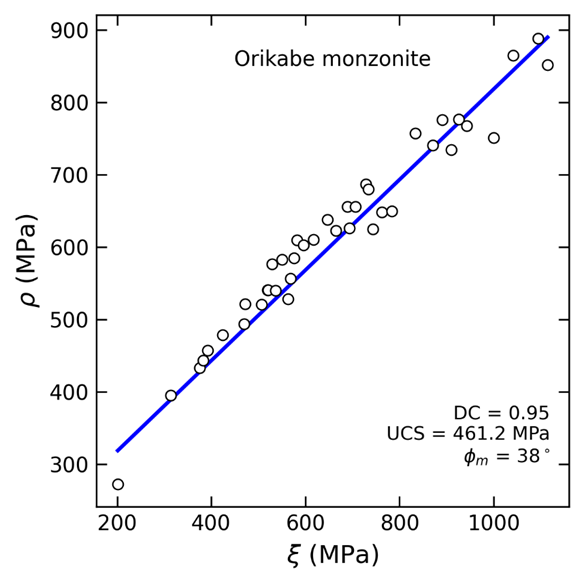}
        \caption{}
        \label{fig:validation-j}
    \end{subfigure}

    \medskip

    \begin{subfigure}[t]{\textwidth}
        \centering
        \includegraphics[width=0.48\textwidth,height=0.26\textheight,keepaspectratio]{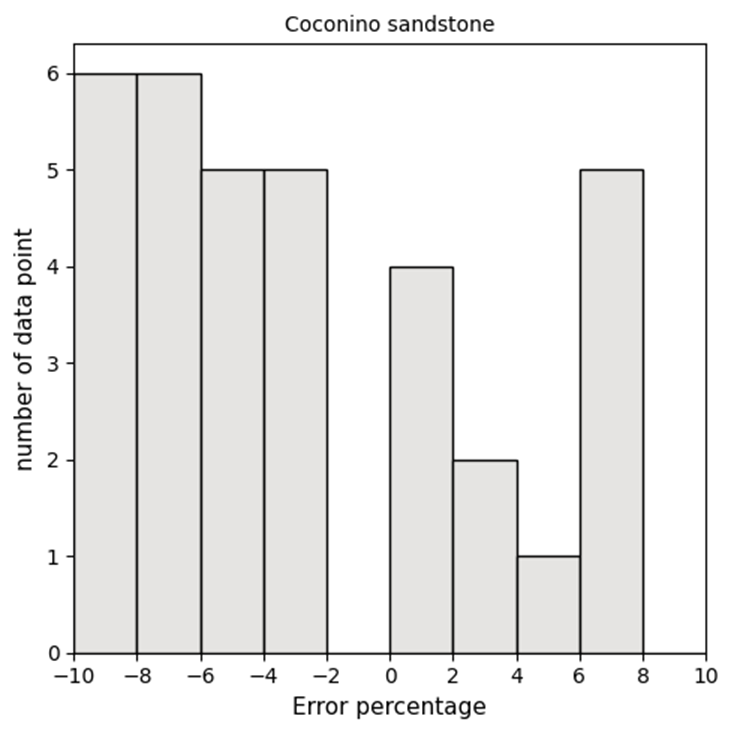}
        \caption{}
        \label{fig:validation-k}
    \end{subfigure}

    \caption[]{(continued).}
\end{figure}
\par
The DC values of $\rho$ for eleven rock types of table(\ref{tab:rock_sources}) are calculated as shown in table (\ref{tab:rmse_dc_rho}). From our result we see that the values of DC for all the eleven rock types are in the range (0.88-0.98). The best ones are for Inada granite (DC=0.98), KTB amphibolite (DC=0.97), Yamaguchi marble (DC=0.96), Orikabe monzonite (DC=0.95) and Manazuru andesite (DC=0.95). The minimum is for Mizuho trachyte (DC=0.88) and for the rest of rock type DC is  greater than 0.90.
\par
\subsection{Root Mean Squared Error (RMSE).}The predictive performance of the proposed failure criterion was further evaluated using the root mean square error (RMSE) calculated in terms of $\rho$ for the eleven rock types according to eqn as
\begin{equation}
    RMSE=\sqrt{\frac{1}{n}\sum_{i}^{n}e_i^2} \quad i=1,2,...,n
    \label{eqn23}
\end{equation}
\par
Where n is the number of experimental data points of a particular rock. $e_i$ is the error defined as 
\begin{equation}
    e_i=\rho_i^{pre}-\rho_i^{exp}
\end{equation}
$\rho_i^{pre}$ is $i^{th}$ predicted value by the proposed model and $\rho_i^{exp}$ is the $i^{th}$ experimental data point. A smaller RMSE denotes a better agreement between predicted and experimental data.
\par
The calculated RMSE values as shown in table (\ref{tab:rmse_dc_rho}) range from 5.13 to 36.62, indicating variations in the prediction accuracy among the different rock types. The lowest RMSE values are obtained for Yuubari shale (5.13) and Shirahama sandstone (9.30).In contrast, the highest RMSE values are observed for KTB amphibolite (36.62) and Inada granite (30.78), indicating relatively larger deviations between the model predictions and experimental results.

\subsection{Percentage Error} The percentage error measures the relative difference between the experimentally measured rock strength and the strength predicted by a failure criterion. It helps evaluate how accurately a model predicts the experimental data. It is defined as,
\begin{equation}
    \epsilon_i=\frac{\text{predicted value - tested value}}{\text{tested value}}*100\%
\end{equation}
\par
Or with respect to $\rho$
\begin{equation}
    \epsilon_i=\frac{\rho_i^{pred}-\rho_i^{test}}{\rho_i^{test}}*100\%
    \label{eqn29}
\end{equation}

\begin{figure}[p]
    \centering

    \begin{subfigure}[t]{0.48\textwidth}
        \centering
        \includegraphics[width=\linewidth,height=0.26\textheight,keepaspectratio]{Error_percen/KTB.png}
        \caption{}
        \label{fig:percentage-error-a}
    \end{subfigure}
    \hfill
    \begin{subfigure}[t]{0.48\textwidth}
        \centering
        \includegraphics[width=\linewidth,height=0.26\textheight,keepaspectratio]{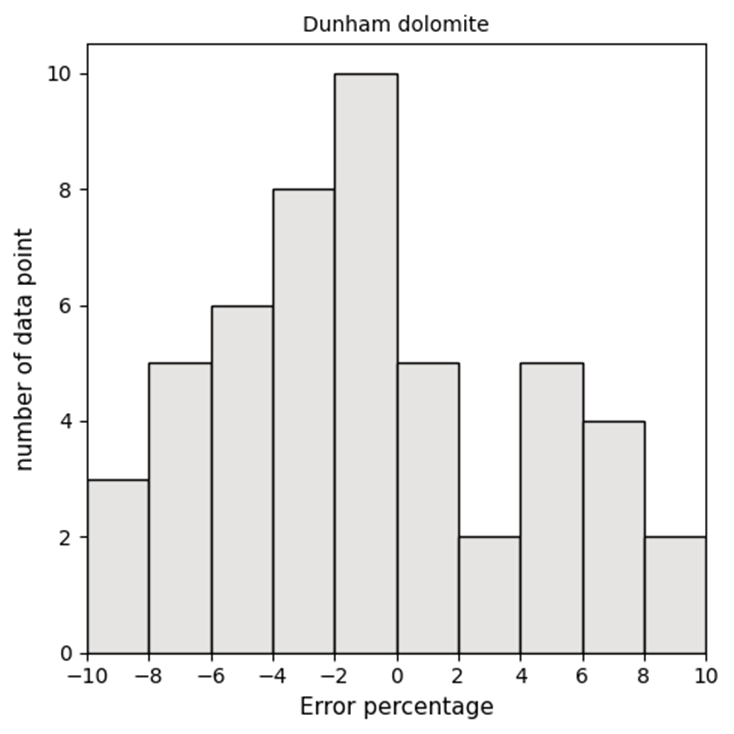}
        \caption{}
        \label{fig:percentage-error-b}
    \end{subfigure}

    \medskip

    \begin{subfigure}[t]{0.48\textwidth}
        \centering
        \includegraphics[width=\linewidth,height=0.26\textheight,keepaspectratio]{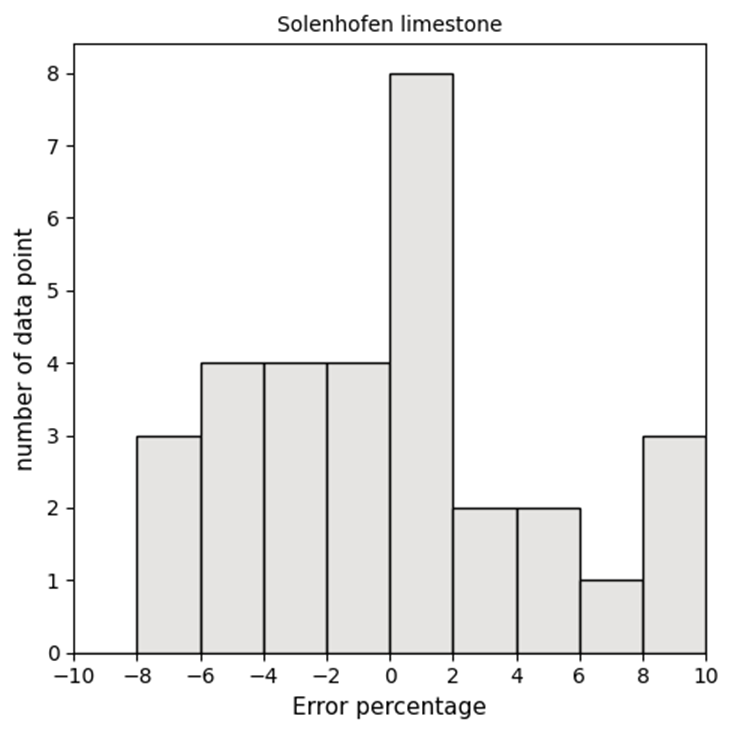}
        \caption{}
        \label{fig:percentage-error-c}
    \end{subfigure}
    \hfill
    \begin{subfigure}[t]{0.48\textwidth}
        \centering
        \includegraphics[width=\linewidth,height=0.26\textheight,keepaspectratio]{Error_percen/shirahama.png}
        \caption{}
        \label{fig:percentage-error-d}
    \end{subfigure}

    \medskip

    \begin{subfigure}[t]{0.48\textwidth}
        \centering
        \includegraphics[width=\linewidth,height=0.26\textheight,keepaspectratio]{Error_percen/yuubari.png}
        \caption{}
        \label{fig:percentage-error-e}
    \end{subfigure}
    \hfill
    \begin{subfigure}[t]{0.48\textwidth}
        \centering
        \includegraphics[width=\linewidth,height=0.26\textheight,keepaspectratio]{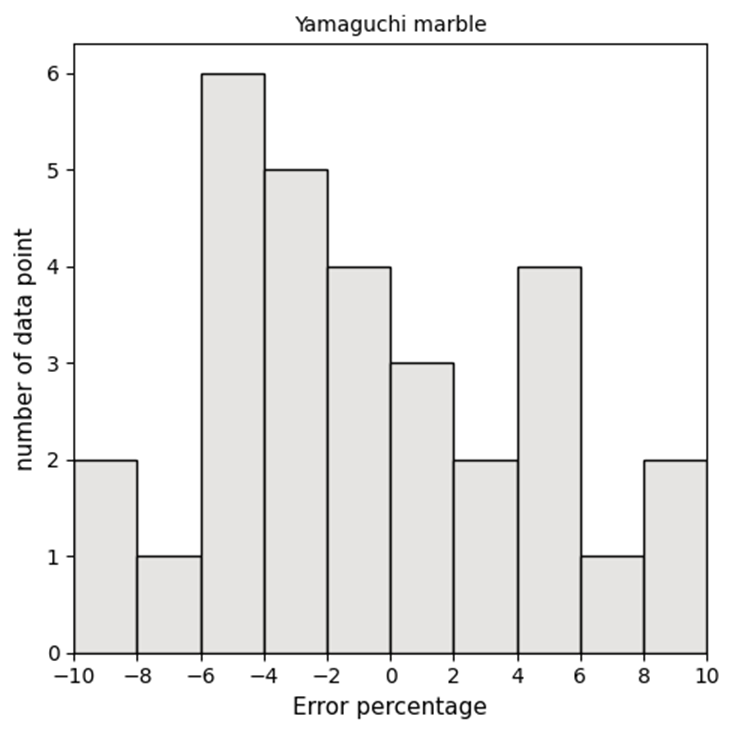}
        \caption{}
        \label{fig:percentage-error-f}
    \end{subfigure}

    \caption{Distribution of percentage misfit between predicted and experimental results for eleven rock types under triaxial compression using the proposed new criterion.}
    \label{fig:percentage-error}
\end{figure}

\begin{figure}[p]
    \ContinuedFloat
    \centering

    \begin{subfigure}[t]{0.48\textwidth}
        \centering
        \includegraphics[width=\linewidth,height=0.26\textheight,keepaspectratio]{Error_percen/Mizuho.png}
        \caption{}
        \label{fig:percentage-error-g}
    \end{subfigure}
    \hfill
    \begin{subfigure}[t]{0.48\textwidth}
        \centering
        \includegraphics[width=\linewidth,height=0.26\textheight,keepaspectratio]{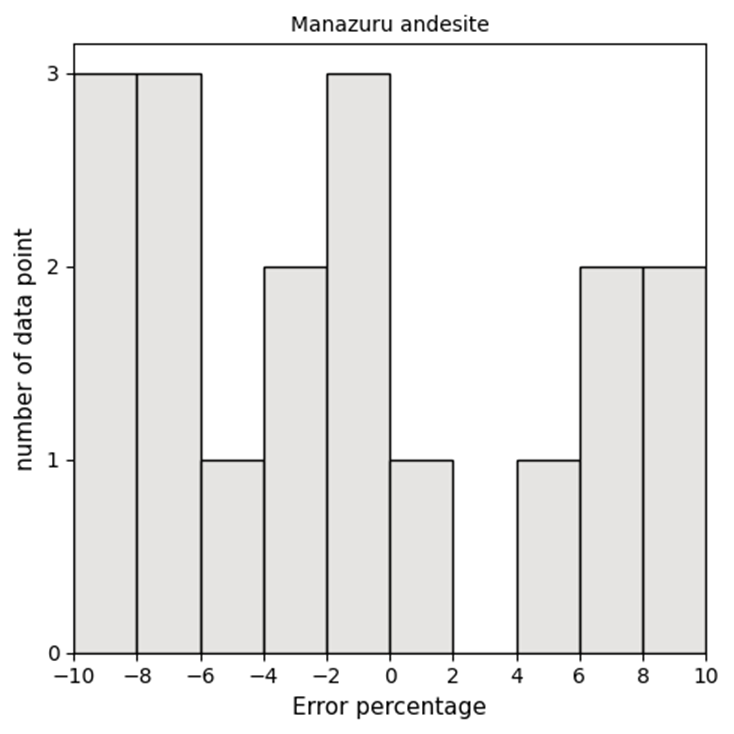}
        \caption{}
        \label{fig:percentage-error-h}
    \end{subfigure}

    \medskip

    \begin{subfigure}[t]{0.48\textwidth}
        \centering
        \includegraphics[width=\linewidth,height=0.26\textheight,keepaspectratio]{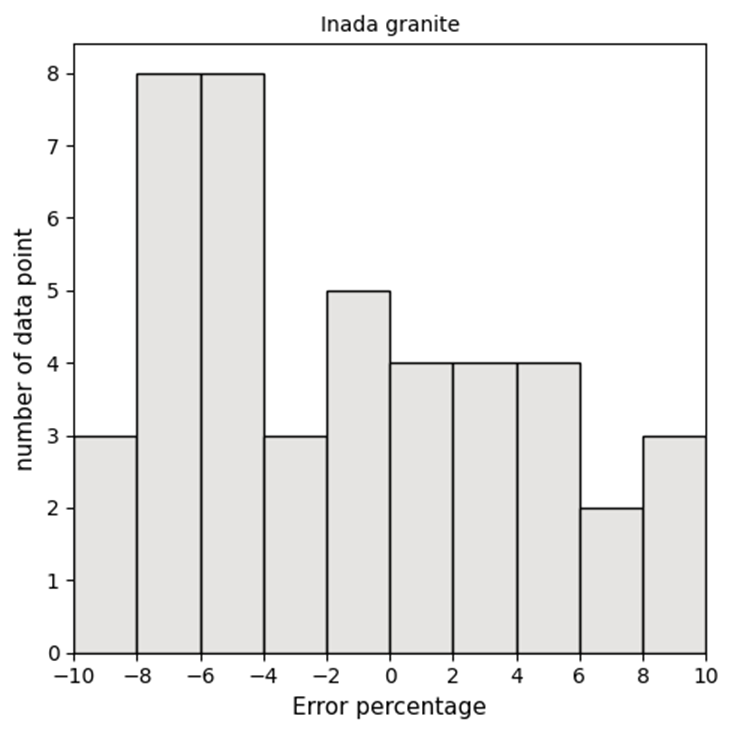}
        \caption{}
        \label{fig:percentage-error-i}
    \end{subfigure}
    \hfill
    \begin{subfigure}[t]{0.48\textwidth}
        \centering
        \includegraphics[width=\linewidth,height=0.26\textheight,keepaspectratio]{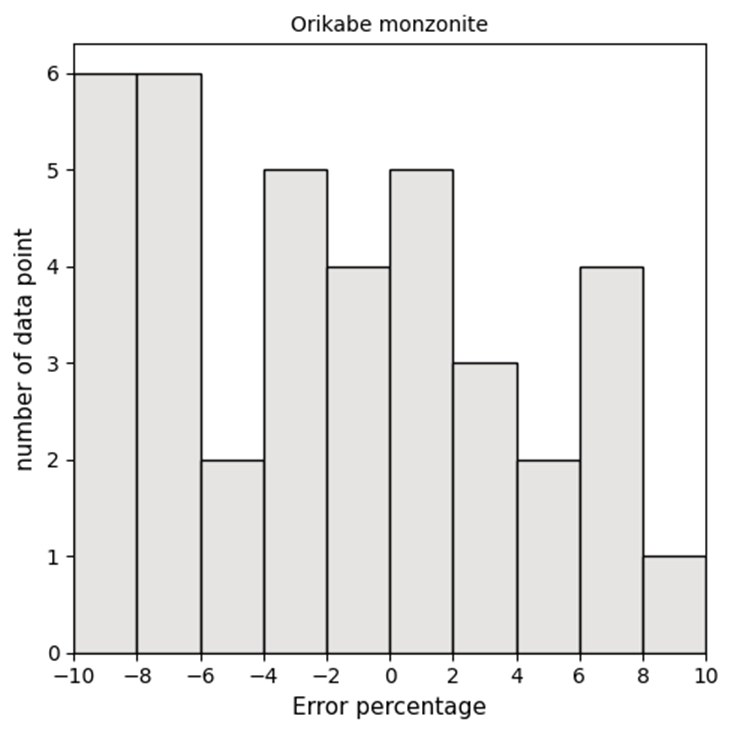}
        \caption{}
        \label{fig:percentage-error-j}
    \end{subfigure}

    \medskip

    \begin{subfigure}[t]{\textwidth}
        \centering
        \includegraphics[width=0.48\textwidth,height=0.26\textheight,keepaspectratio]{Error_percen/coconino.png}
        \caption{}
        \label{fig:percentage-error-k}
    \end{subfigure}

    \caption[]{(continued).}
\end{figure}
\par
Here, $\rho_i^{pred}$ and $\rho_i^{test}$ are the $i^{th}$ predicted and tested value values of $\rho$ on the triaxial test data. $\epsilon >0$ indicates an overestimation of the rock strength and $\epsilon<0$ indicates an underestimation of the rock strength.
\par
Fig.(\ref{fig:percentage-error}) shows the percentage error distribution for the predicted $\rho$ for all the mentioned rock types.The percentage errors are calculated using the triaxial compression test data for all eleven rock types, as defined in Eqn. (\ref{eqn29}). Figure 4 illustrates the distribution of percentage errors, where the x-axis represents the percentage error intervals and the y-axis indicates the corresponding number of experimental data points within each interval. The percentage error distributions show that the predicted values are within the range of approximately $-10\%$ to $+10\%$ for most rock types, while KTB amphibolite exhibits a comparatively wider distribution extending from $-15\%$ to $+15\%$.
\section{Comparasion of the New Generalised criterion with some existing 3D failure criteria.}To evaluate the effectiveness and predictive capability of the proposed new generalized failure criterion, a comparative analysis is performed with some existing 3D failure criteria namely, New Extended Mohr-Coulomb, Mogi-Coulomb and Drucker-Prager.
\par
These criteria are selected to assess the ability of the proposed model to represent the failure behaviour of different rock types under complex stress conditions, particularly considering the influence of the intermediate principal stress ($\sigma_2$). The comparison is carried out using experimental triaxial compression test data for eleven rock types. The predictive performance of each criterion is evaluated in terms of root mean square error (RMSE) and determination coefficient (DC) of ($\rho$) as shown in table(\ref{tab:rmse_dc_rho}). 
\par
This comparative assessment aims to examine the accuracy and applicability of the proposed criterion relative to the existing three-dimensional failure criteria.

\subsection{Comparison with New Extended Mohr-Coulomb failure criterion. }A New Extended Mohr-Coulomb failure criterion \cite{mahetaji2023new} was developed in 2023 by Mahentaji by considering the weighting average of principal stresses.
\par
It introduces the effect of intermediate principal stress as a weighting P of $\sigma_2$ and the weighting of $\sigma_1$ and $\sigma_3$ are equal to unity. It is express as given \cite{mahetaji2023new}
\begin{equation}
    \sigma_1=\sigma_2q_1+\sigma_3q_2+C.S
\end{equation}
Here,
\begin{align*}
q_1 &= \frac{2P\sin\phi_m}
{(2-2\sin\phi_m+P)},\\
q_2 &= \frac{(2+2\sin\phi_m+P)}
{(2-2\sin\phi_m+P)},\\
CS &= \frac{2(2+P)c\cos\phi_m}
{(2-2\sin\phi_m+P)}.
\end{align*}
\par
c is the cohesion, $\phi_m$ is the 3-dimensional friction angle and P is the weighting of $\sigma_2$. It is writen in Haigh-Westergaard coordinates as given,
\begin{equation}
    \rho=\frac{\sqrt{2}(q_1+q_2-1)\xi+\sqrt{3}CS)}{(\sin\theta +\sqrt{3}\cos\theta)q_2 - (2\sin\theta)q_1-\sin\theta +\sqrt{3}\cos\theta}
\end{equation}
\begin{figure}[htbp]
    \centering
    \includegraphics[width=\textwidth]{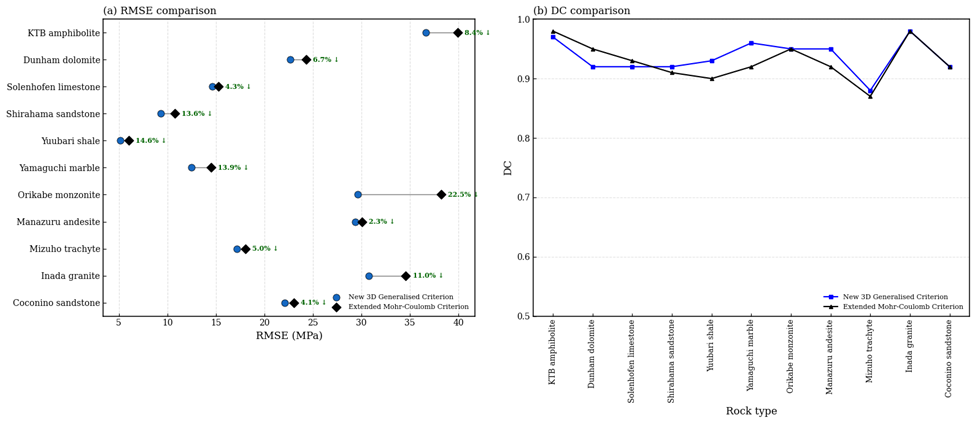}
    \caption{Comparison of the New Generalised (NG) and New Extended (NE) criteria.}
    \label{fig:ng-vs-ne}
\end{figure}
\par
The DC and RMSE (MPa) values of $\rho$ for eleven rock triaxial data points are calculated for the New Extended Mohr--Coulomb criterion and compared with our proposed New Generalised criterion, as shown in Figure~\ref{fig:ng-vs-ne}.
\par
The RMSE values indicate that the proposed 3D Generalised criterion provides comparatively better predictions than the New Extended Mohr–Coulomb criterion. The proposed criterion exhibits a maximum RMSE of 36.62 for KTB Amphibolite, compared with 39.96 for the New Extended Mohr–Coulomb criterion. Similarly, the minimum RMSE is 5.13 for Yuubari Shale using the proposed criterion, whereas the corresponding value for the New Extended Mohr-Coulomb criterion is 6.01. For the remaining rock types, the proposed criterion also demonstrates slightly lower RMSE values, indicating an overall improvement in predictive accuracy compared with the New Extended Mohr–Coulomb criterion.
\par
The desirable value of the determination coefficient (DC) is close to 1, within the range of 0 to 1. For the proposed criterion, the highest DC values are obtained for Inada Granite (0.98) and Dunham Dolomite (0.92), while the lowest value is observed for Mizuho Trachyte (0.88). In comparison, the New Extended Mohr–Coulomb criterion exhibits the highest DC values for Inada Granite (0.98) and KTB Amphibolite (0.98), with the lowest value again observed for Mizuho Trachyte (0.87). For the remaining rock types, both criteria demonstrate comparable and satisfactory DC values, all exceeding 0.90.

\subsection{Comparison with the Mogi-Coulomb failure criterion.}
The Mogi–Coulomb criterion, developed by combining Mogi's \cite{mogi1971fracture} failure criterion with the classical Mohr–Coulomb criterion, considers the octahedral shear stress ($\tau_{oct}$) as a function of the mean effective stress along the ($\sigma_2)$ direction. The criterion was formulated by Al-Ajmi and Zimmerman \cite{al2005relation} and is mathematically expressed as:
\begin{equation}
    \tau_{oct}=a+b \sigma_{m,2}
\end{equation}
\par
a and b are the material constants determined from experimental data. And it's equivalent relation with cohesion and friction angle as,
$$a=\frac{2\sqrt{2}}{2}c\cos\phi$$
$$b=\frac{2\sqrt{2}}{3}\sin\phi$$
And $\tau_{oct}=\frac{\sqrt{(\sigma_1-\sigma_2)^2+(\sigma_2-\sigma_3)^2+(\sigma_3-\sigma_1)^2}}{3}, \sigma_{m,2}=\frac{\sigma_1+\sigma_3}{2}$
It is written in Haigh-Westergaard coordinates as
\begin{equation}
    \rho=\sqrt{3}a+\frac{b}{2} (3\xi-\sqrt{3} b)
\end{equation}
\begin{figure}[htbp]
    \centering
    \includegraphics[width=\textwidth]{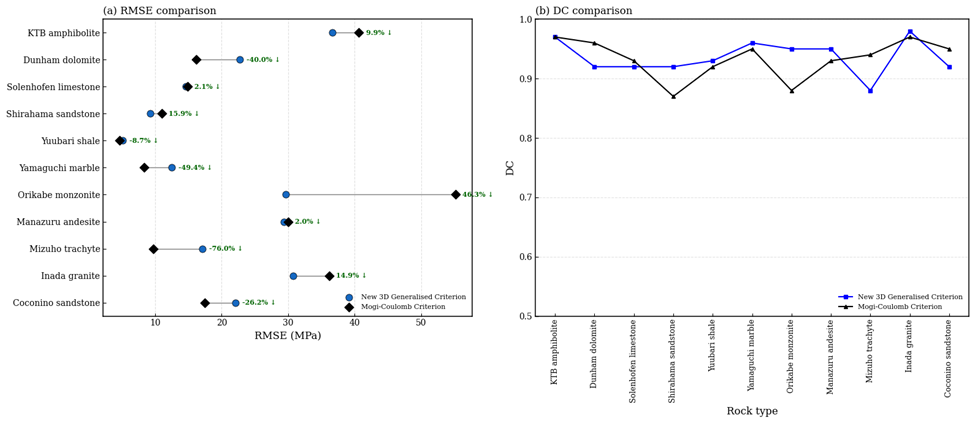}
    \caption{Comparison of the New Generalised (NG) and Mogi-Coulomb(MG) criteria.}
    \label{fig:ng-vs-mg}
\end{figure}
\par
The comparison between the New 3D Generalized Criteria and the Mogi-Coulomb Criteria is made following the same procedure described in Subsection 6.1, and the corresponding results are presented in Figure \ref{fig:ng-vs-mg}. It is observed that the RMSE values of the Mogi–Coulomb criterion range from 4.72 MPa to 55.15 MPa for Yuubari shale and Orikabe monzonite, respectively. In comparison, the RMSE values of the proposed New 3D Generalised criterion range from 5.13 MPa to 36.62 MPa for Yuubari shale and KTB amphibolite, respectively. Among the eleven rock types considered, the proposed criterion exhibits lower RMSE values for six rock types, namely KTB amphibolite, Orikabe monzonite, Solnhofen limestone, Shirahama sandstone, Manazuru andesite, and Inada granite. For the remaining five rock types, namely Mizuho trachyte, Yuubari shale, Yamaguchi marble, Coconino sandstone, and Dunham dolomite, both criteria demonstrate comparable RMSE values.

Regarding the Coefficient of Determination (DC), the proposed New 3D Generalised criterion exhibits better performance for six rock types, namely Shirahama sandstone, Inada granite, Manazuru andesite, Yamaguchi marble, Yuubari shale, and Orikabe monzonite. For the remaining five rock types, the DC values obtained from both criteria are comparable. 
\par
Overall, the comparison indicates that the proposed criterion provides comparable or improved predictive performance relative to the Mogi–Coulomb criterion in terms of both RMSE and DC.

\subsection{Comparision with Drucker-Prager failure criterion.}
The Drucker–Prager (DP) criterion, originally proposed by Drucker and Prager \cite{drucker1952soil}, is a widely used pressure-dependent failure criterion in rock and soil mechanics. 
\par
It represents the failure surface as a circular cone in the principal stress space and is commonly considered a smooth approximation of the Mohr–Coulomb failure criterion.  Its mathematical expression can be written as
\begin{equation}
    \tau_{oct}= \sqrt{\frac{2}{3}}(3\lambda \sigma_{oct}+k)
\end{equation}
\par
$\tau_{oct}$ is the octahedral shear stress and $\sigma_{oct}$ is the octahedral stress, $\lambda$ and $k$ are material parameters and can be related to Mohr-Coulomb parameters \cite{alejano2014drucker} 
\begin{equation}
    \lambda=\frac{2\sin\phi}{\sqrt{3}(3-\sin\phi)}
\end{equation}
\begin{equation}
    k=\frac{6c\cos\phi}{\sqrt{3}(3-\sin\phi)}
\end{equation}
\par
It is expressed in Haigh-Westergaard coordinates $(\rho, \xi, \theta)$ 
\begin{equation}
    \rho=\sqrt{2}k+\sqrt{6}\lambda \xi
\end{equation}
\begin{figure}[htbp]
    \centering
    \includegraphics[width=\textwidth]{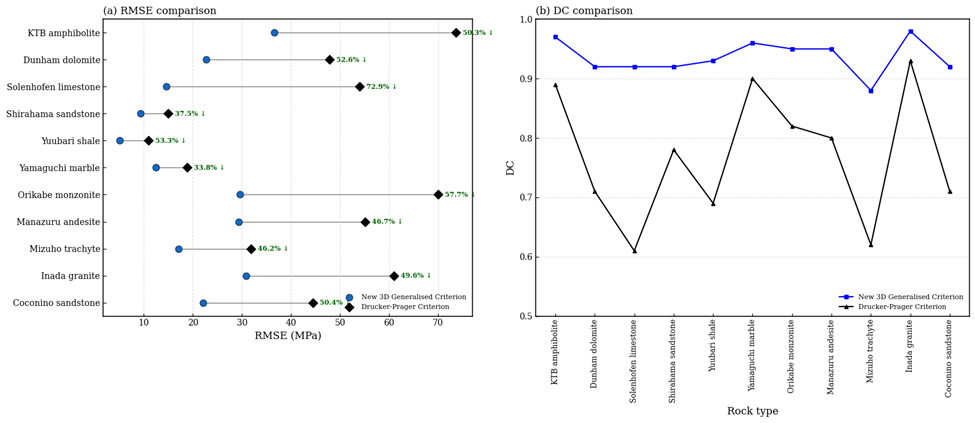}
    \caption{Comparison of the New Generalised (NG) and Drucker-Prager(DP) criteria.}
    \label{fig:ng-vs-dp}
\end{figure}
\par
\noindent The same procedure is followed to compare the proposed New 3D Generalised criterion with the Drucker–Prager criterion. As shown in figure \ref{fig:ng-vs-dp}, the Drucker-Prager criterion exhibits comparatively higher RMSE values and lower DC values than the proposed criterion for all eleven rock types. These results suggest that the proposed criterion provides better agreement with the experimental data and lower prediction errors than the Drucker–Prager criterion for the rock types considered.
\section{Comparative Discussion of Failure Criteria}
The proposed New Generalised 3D failure criterion was calibrated using triaxial compression test data from eleven different rock types. The calibrated parameters, including the intermediate principal stress weighting coefficient (K), friction angle ($\phi)$, and cohesion (c), are presented in Table 2. The calibrated values of (K) range from 0.03 to 0.38, indicating variations in the contribution of the intermediate principal stress among the investigated rock types.
\par
To examine the predictive capability of the proposed criterion in three-dimensional stress space, the calibrated parameters are subsequently used to transform the failure criterion into Haigh–Westergaard coordinates $(\xi,\rho,\theta)$. The validation results for the eleven rock types are presented in Figure (\ref{fig:validation-xi-rho}). The experimental data points exhibit a generally close agreement with the predicted failure envelopes, indicating that the proposed formulation can represent the relationship between the hydrostatic and deviatoric components of the stress state.
\par
The predictive performance of the proposed New Generalised 3D (NG) criterion is compared with three existing three-dimensional failure criteria: the New Extended Mohr–Coulomb (NE), Mogi–Coulomb (MG), and Drucker–Prager (DP) criteria. The comparison is conducted using the same eleven rock types and is evaluated in terms of RMSE and DC of the deviatoric stress parameter ($\rho$). The numerical results are summarised in Table 3 and illustrated in Figures 9-10.

\begin{figure}[htbp]
    \centering
    \includegraphics[width=\textwidth]{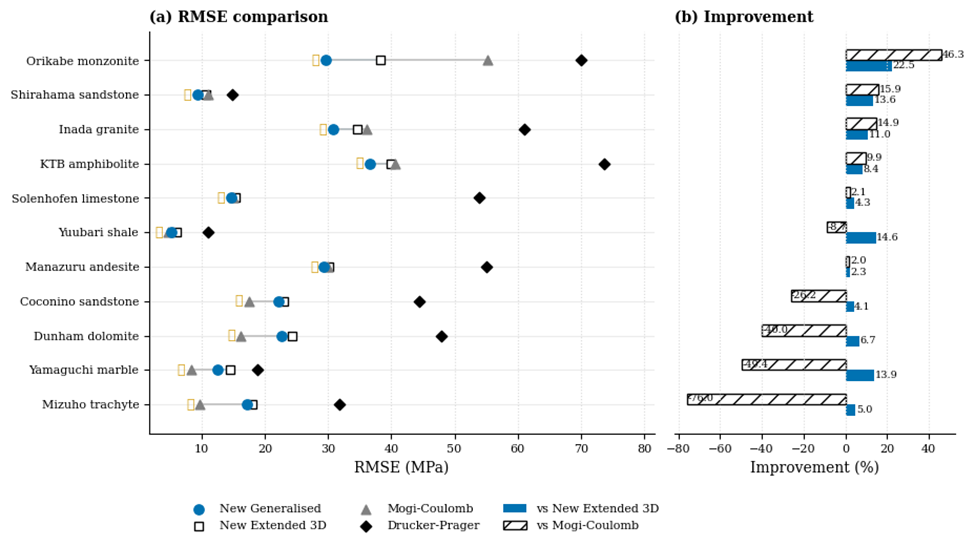}
    \caption{Comparison of RMSE of ($\rho$) of the New Generalised (NG) criterion and other existing 3D criteria.}
    \label{fig:Compare1}
\end{figure}
\begin{figure}[htbp]
    \centering
    \includegraphics[width=\textwidth]{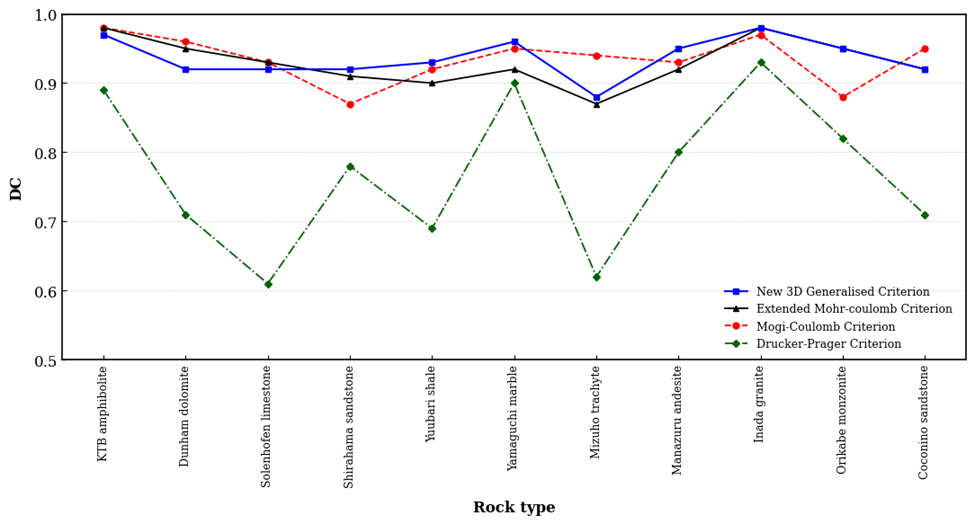}
    \caption{Comparison of DC of ($\rho$) of the New Generalised (NG) criterion and other existing 3D criteria.}
    \label{fig:Compare2}
\end{figure}
\par
We observed that the proposed criterion produces lower RMSE values and higher DC values than Drucker–Prager criterion for all eleven rock types. The results suggest that our formulation provides a closer representation of the experimental data than the Drucker–Prager criterion for the eleven rock types investigated.
\begin{figure}[htbp]
    \centering
    \includegraphics[width=\textwidth]{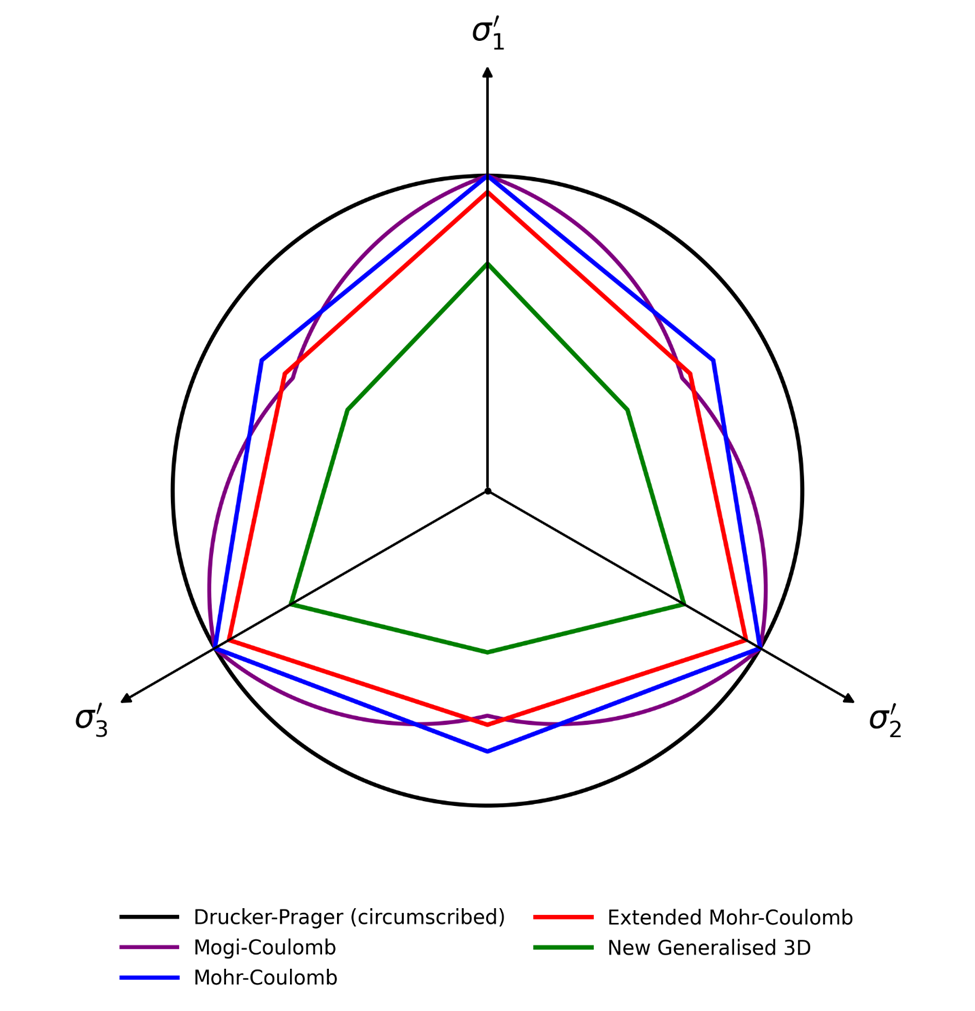}
    \caption{ Comparison of the new 3D criterion with 
existing criteria on the deviatoric plane }
    \label{fig:Compare3}
\end{figure}
\par
The proposed criterion also produces lower RMSE values for all eleven rock than New Extended 3D criterion. The RMSE values of the proposed criterion range from 5.13 to 36.62 MPa, whereas those of the NE criterion range from 6.01 to 39.96 MPa. This indicates a consistent reduction in prediction error for the investigated datasets.
In terms of DC, both criteria demonstrate comparable performance, with values generally exceeding 0.90. The New Generalised 3D criterion produces higher DC values for several rock types, whereas the New Extended 3D criterion exhibits slightly higher values for KTB amphibolite, Dunham dolomite, and Solenhofen limestone. For the remaining rock types, the differences are either small or negligible. These results suggest that introducing unequal weighting contributions of the three principal stresses provides a useful alternative to the weighting formulation adopted in the NE criterion, particularly in reducing the prediction error across the investigated rock types.
\par
The NG criterion exhibits lower RMSE  values than Mogi-Coulomb criterion for six rock types, namely KTB amphibolite, Solenhofen limestone, Shirahama sandstone, Manazuru andesite, Inada granite, and Orikabe monzonite. For the remaining five rock types, the Mogi–Coulomb criterion provides lower RMSE values.
\par
DC comparison indicates that the proposed criterion produces higher values than Mogi-Coulomb criterion for six rock types: Shirahama sandstone, Yuubari shale, Yamaguchi marble, Manazuru andesite, Inada granite, and Orikabe monzonite. For the remaining rock types, the Mogi–Coulomb criterion exhibits either higher or comparable DC values. These observations indicate that the predictive performance of the two criteria varies with rock type. Therefore, the proposed criterion demonstrates comparable or improved predictive capability relative to the Mogi–Coulomb criterion for the experimental datasets considered, rather than uniformly outperforming it.
\par
The prediction error is further examined through the percentage error distributions presented in Figure \ref{fig:percentage-error}. The errors for most rock types are concentrated approximately within the range of $-10\%$ to $+10\%$, while KTB amphibolite exhibits a comparatively wider distribution extending from approximately $-15\%$ to $+15\%$. The distributions provide additional evidence of the agreement between the predicted and experimental results, although the presence of both positive and negative errors indicates that the model does not exclusively overestimate or underestimate the experimental strength.
\par
The deviatoric-plane representation of the proposed criterion is also compared with those of the existing three-dimensional failure criteria, as illustrated in Fig. (\ref{fig:Compare3}). The influence of the intermediate principal stress, governed by \(K=0.1\), is represented by the green solid line, whereas the Extended Mohr–Coulomb criterion, formulated with \(P=0.2\), is represented by the orange solid line. The proposed criterion exhibits a distinct shear-failure envelope, enclosing the smallest volume among the criteria considered. This characteristic arises from the incorporation of the intermediate principal stress through its contribution to the effective normal stress. The resulting smaller failure envelope indicates a more conservative prediction of rock strength under three-dimensional stress conditions, suggesting that the proposed criterion provides a comparatively safer assessment of failure than the other criteria considered.
\section{Conclusion} The present study proposes a new three-dimensional failure criterion that incorporates the influence of the intermediate principal stress ($\sigma_2)$ through unequal weighting coefficients $W_1$, $K$, and $W_3 $ corresponding to the principal stresses ($\sigma_1)$, ($\sigma_2)$, and ($\sigma_3)$, respectively. In the proposed formulation, the contribution of the intermediate principal stress is introduced directly into the effective normal stress, allowing the individual contributions of the three principal stresses to be represented through independent weighting coefficients. 
\par
The proposed criterion retains a simple mathematical formulation and is easy to implement for predicting the failure strength of intact rocks. Furthermore, when the influence of the intermediate principal stress is neglected ($K=0$), the proposed criterion reduces to the conventional Mohr–Coulomb failure criterion, demonstrating its consistency with the established two-dimensional formulation.
\par
The proposed criterion was calibrated using triaxial compression test data from eleven different rock types. The calibrated values of the intermediate principal stress weighting parameter (K) range from 0.03 to 0.38, indicating that the contribution of ($\sigma_2$) varies considerably among different rock materials. The calibrated parameters were subsequently employed to evaluate the proposed criterion in the Haigh–Westergaard coordinate system. The validation results demonstrate a generally good agreement between the predicted failure envelopes and the experimental data. The coefficient of determination (DC) values range from 0.88 to 0.98, while the corresponding RMSE values range from 5.13 to 36.62 MPa. Furthermore, the percentage error distributions indicate that the prediction errors for most rock types are concentrated approximately within the range of $-10\%$ to $+10\%$, demonstrating the capability of the proposed criterion to reproduce the experimental failure behaviour.
\par
To understand the failure phenomenon, Extended Mohr-Coulomb, Drucker-Prager and Mogi-Coulomb were also run on the same triaxial compression datasets. The result shows that the proposed criterion consistently produces lower RMSE values than the New Extended Mohr–Coulomb and Drucker–Prager criteria for all eleven investigated rock types. In comparison with the New Extended Mohr–Coulomb criterion, the RMSE values decrease from a range of 6.01–39.96 MPa to 5.13–36.62 MPa. The DC values of both criteria remain generally comparable, with most values exceeding 0.90. In comparison with the Mogi–Coulomb criterion, the proposed formulation exhibits lower RMSE values for six rock types and higher DC values for six rock types, indicating comparable or improved predictive performance depending on the rock material. These findings suggest that the proposed formulation provides a useful and simpler  formulation alternative to existing three-dimensional failure criteria, particularly in terms of reducing prediction errors across the investigated datasets.
\par
The deviatoric-plane comparison further illustrates the influence of the intermediate principal stress on the geometry of the proposed failure envelope. The proposed criterion encloses a comparatively smaller failure envelope, indicating a safer prediction of rock strength.
\par
Overall, the proposed generalized 3D failure criterion provides a simple and computationally convenient framework for incorporating the unequal contributions of the three principal stresses into rock failure prediction. The formulation may provide a useful basis for further developments in wellbore stability analysis under complex in-situ stress conditions. Future investigations should focus on extending the formulation to account for rock anisotropy and damage evolution. Such developments would help establish its applicability to more complex geological and subsurface engineering conditions.
\bibliographystyle{plainnat}
\bibliography{Reference}
\end{document}